%% file: acl_latex.tex
\documentclass[11pt]{article}
\usepackage[]{acl}
\usepackage{times}
\usepackage{latexsym}

\usepackage[T1]{fontenc}

\usepackage[utf8]{inputenc}

\usepackage{microtype}
\usepackage{inconsolata}
\usepackage{graphicx}
\usepackage{algorithm}
\usepackage{algpseudocode}
\usepackage{amsmath}
\usepackage{booktabs}
\usepackage{amssymb}
\usepackage{multirow}
\usepackage{xcolor}
\usepackage{algorithm}
\usepackage{enumitem}
\usepackage{subcaption}
\usepackage{makecell}
\usepackage{xcolor}
\usepackage{hyperref}

\usepackage[most]{tcolorbox}
\usepackage{listings}
\usepackage{cuted}
\usepackage{placeins} 
\usepackage{stfloats}

\title{Compress to Focus: Efficient Coordinate Compression for Policy Optimization in Multi-Turn GUI Agents}

\author{%
  \bfseries
  Yurun Song\textsuperscript{1,2,*} \;
  Jiong Yin\textsuperscript{1,3,*} \; 
  Rongjunchen Zhang\textsuperscript{1,$\spadesuit$}\;
  Ian G. Harris\textsuperscript{2} \;
\\[1.2ex]
  \small\itshape
  \textsuperscript{1}HiThink Research \;
  \textsuperscript{2}University of California, Irvine \;
  \textsuperscript{3}Hangzhou Dianzi University \\[1.2ex]
  {
  \ttfamily\small
  {\{yuruns,iharris\}@uci.edu} \;
  {jiong.yin@hdu.edu.cn} \;
  {zhangrongjunchen@myhexin.com}
  }
}

\begin{document}
\maketitle
\begin{abstract}

Multi-turn GUI agents enable complex task completion through sequential decision-making, but suffer from severe context inflation as interaction history accumulates. 
Existing strategies either sacrifice long-term context via truncation or compromise spatial structure through token pruning.
In this paper, we propose \textbf{C}oordinate \textbf{C}ompression \textbf{P}olicy \textbf{O}ptimization~\textbf{(CCPO)}, an efficient policy optimization framework that couples visual compression with policy optimization for multi-turn GUI agents. 
CCPO introduces \textbf{C}oordinate-\textbf{A}ware \textbf{S}patial Compression~\textbf{(CASC)}, which aggregates coordinates from multiple rollouts to capture target-relevant regions and progressively narrow historical attention around key visual areas. From interactions across rollouts, CASC adaptively constructs attention boundaries that concentrate computation on the most informative regions of the scene.
We further design a Distance-Based Advantage that provides fine-grained learning signals based on distance rather than binary correctness, improving both grounding accuracy and compression quality. 
Extensive experiments demonstrate that CCPO achieves SOTA performance across four benchmarks with up to 55\% token compression and 3.8$\times$ training speedup. Our code can be available at~\href{https://hithink-research.github.io/CCPO/}{https://hithink-research.github.io/CCPO}
\end{abstract}


\input{sec/introduction}

\begin{figure*}[]
  \vspace{-1.5em}
  \includegraphics[width=\textwidth]{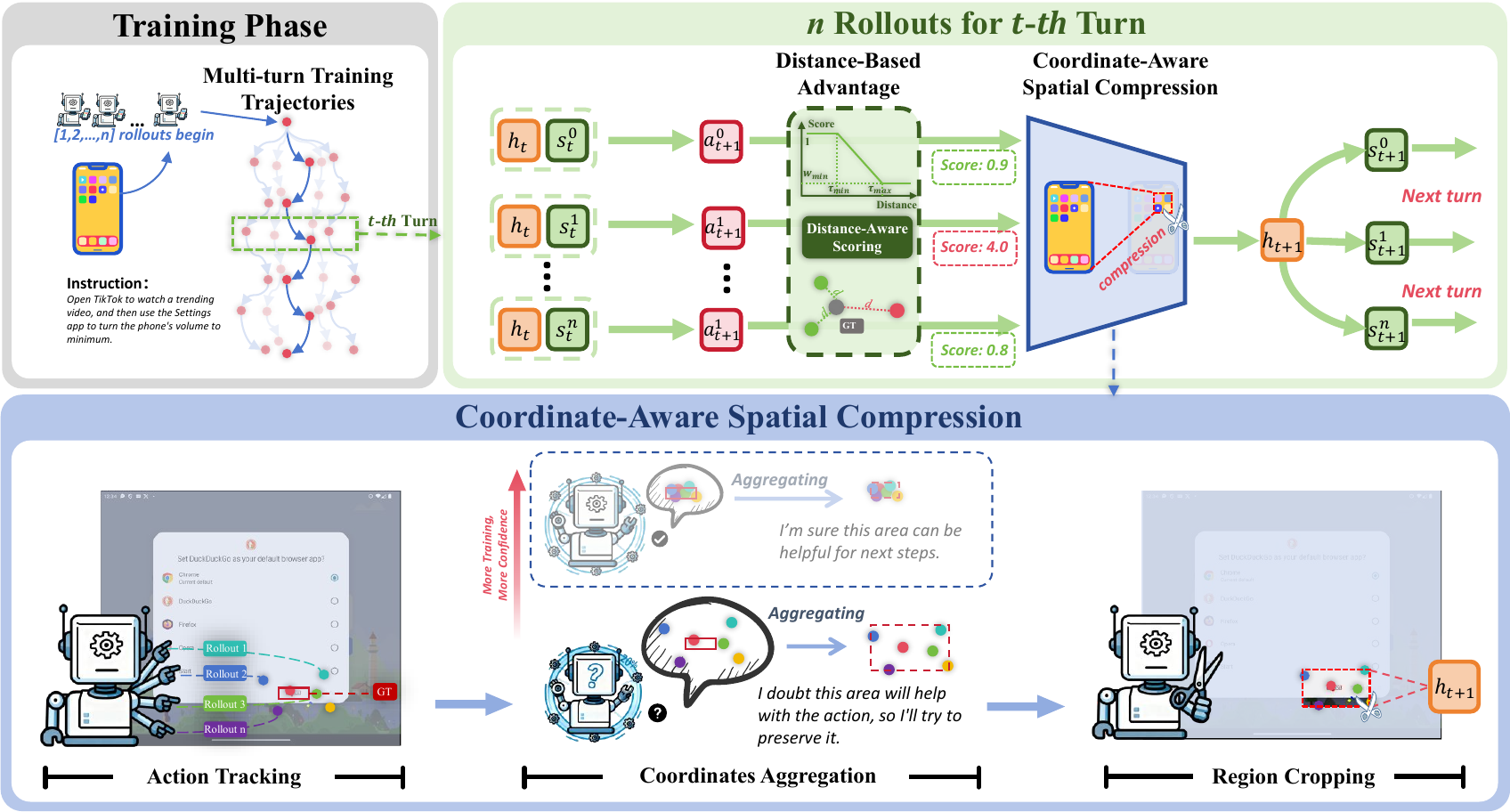} 
  \caption {
  Overview of CCPO framework. 
  The training phase~(top) optimizes policies via multi-turn rollouts evaluated by the \textbf{Distance-Based Advantage}. 
  The \textbf{Coordinate-Aware Spatial Compression} module~(bottom) tracks $n$ actions and aggregates coordinates to predict ROI of each step, then crops the task-relevant region as a focused visual history $h_{t+1}$.
  }\label{framework}
\end{figure*}

\input{sec/related_work}

\section{Method}

\input{sec/method}

\section{Experiments}
\input{sec/experiment}

\section{Main Results}

\input{sec/results}

\section{Analysis}
\input{sec/analysis}

\section{Conclusion}

\input{sec/conclusion}

\clearpage
\bibliography{custom}

\clearpage
\appendix
\section{Appendix}
\label{sec:appendix}
\input{sec/appendix}

\end{document}

%% file: sec/introduction.tex
\section{Introduction}
\begingroup
\renewcommand\thefootnote{}%
\setlength{\parindent}{0pt}%
\footnotetext{* Equal contribution. Intern at HiThink Research.}
\footnotetext{$\spadesuit$ Corresponding Author}
\addtocounter{footnote}{-2}%
\endgroup


GUI automation enables agents to execute sophisticated tasks by capturing multimodal cues.
However, complex real-world workflows render single-turn interaction inadequately.
Thus, effective automation requires multi-turn capabilities to execute precise decisions based on historical context.

Although multi-turn interaction facilitates complex tasks,
existing GUI agents~\cite{lu2025guiodyssey,chen2025less,cheng2024seeclick} struggle with context inflation. 
Therefore, the accumulation poses two critical challenges.
First, computational costs increase rapidly as context length grows. In complex scenarios, contexts can easily exceed 32k tokens, which imposes substantial memory and latency burdens. 
Second, not all historical information contributes equally to current decisions.
Visual tokens often contain redundancy, while naive truncation discards critical spatial information.
Together, these issues motivate a selective compression approach that preserves task-critical information while filtering out irrelevant context.

\begin{figure}[tp]
    \centering
    \setlength{\abovecaptionskip}{0.01cm}
    \setlength{\belowcaptionskip}{0.01cm}
    \includegraphics[width=\columnwidth]{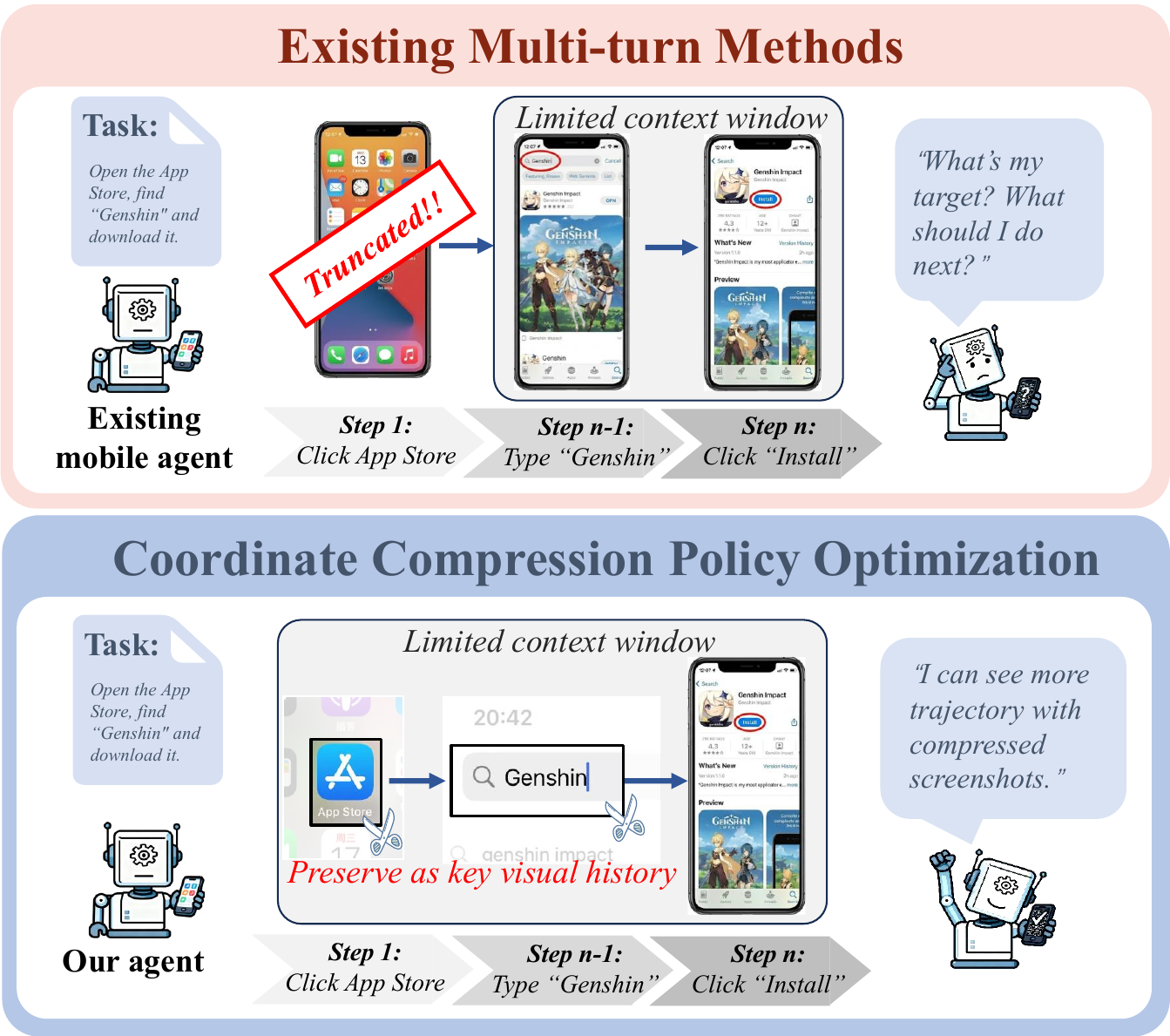} 
    \caption{
    \textbf{Top:} Existing multi-turn methods tend to truncate the visual history due to the limited context length. 
    \textbf{Bottom:} CCPO preserves the key visual history to maintain the longer trajectory visibility.
    }
    \vspace{-0.6cm}
    \label{fig_intro}
\end{figure}

To mitigate the context inflation issue, existing methods~\cite{cheng2024seeclick,chen2025less} employ two strategies, yet both exhibit fundamental limitations in GUI scenarios. 
Direct truncation~\cite{cheng2024seeclick,lu2025guiodyssey,lin2024showui} aims to preserve all visual information, but it is tightly constrained by the context window, which prevents tracking long-range dependencies.
Conversely, token pruning~\cite{chen2025less} summarizes visual context but fails to explicitly represent historical trajectories.
This method disrupts the correspondence between actions and their spatial information, introducing ambiguity that hinders precise localization.
Therefore, GUI agent compression faces two critical challenges:
\textbf{1) Mismatch between spatial locality and temporal dependency.} 
While agents require long-term history for task coherence, 
the visual cues relevant to each decision are inherently localized around the regions where actions occur.
This suggests that action trajectories should be preserved across turns, 
%
whereas retaining the full screen at every step is largely redundant.
\textbf{2) Coupled optimization of compression and action policies.}
Unlike static compression methods, 
a dynamic compression method requires foreknowledge of action-relevant regions, while accurate action prediction depends on well-compressed context. 
This bidirectional dependency leads to a training dilemma, as 
optimizing either objective alone often leads to suboptimal convergence.

In light of these insights, we propose \textbf{C}oordinate \textbf{C}ompression \textbf{P}olicy \textbf{O}ptimization~(CCPO), an efficient multi-turn policy optimization framework to couple visual compression with policy learning.
The key insight of CCPO is that (i) task-relevant visual cues are spatially localized to a few critical regions, and (ii) temporal coherence is maintained by continuously tracking and preserving the trajectories of these regions over time.
To this end, we introduce Coordinate-Aware Spatial Compression (CASC) to dynamically narrow attention boundaries with various action prediction.
Specifically, CASC tracks interaction coordinates from predicted actions and aggregates them to compute the region of interest (ROI), then crops the visual context accordingly. 
This forms a virtuous cycle where improved visual focus yields better coordinate predictions, progressively tightening the spatial boundaries.
Furthermore, we design the Distance-Based Advantage to replace binary feedback with smooth, distance-based supervision, thereby guiding the policy to progressively converge toward precise target locations.


\noindent\textbf{Our contributions can be summarized as:}
\begin{itemize}[leftmargin=*,nolistsep]
    \item We propose CCPO, a unified reinforcement learning framework where compression and coordinate prediction are optimized in a beneficial loop.
    \item We introduce Coordinate-Aware Spatial Compression, which dynamically constructs spatial attention boundaries from interaction trajectories, achieving up to 55\% token reduction and 3.8$\times$ training speedup.
    \item We design a Distance-Based Advantage that provides soft distance-based guidance for coordinate-related actions to improve prediction accuracy and grounding abilities.
    \item Extensive experiments show our method achieves the state-of-the-art results on four public datasets.
\end{itemize}

%% file: sec/related_work.tex
\section{Related Works}

\noindent
\textbf{GUI Agents with Reinforcement Learning.}
Recent progress in GUI automation has been predominantly shaped by two distinct paradigms~\cite{survey_1, survey_2, survey_3, survey_4, wang2025opencua, zhang2025appagent}.
The first generation of methods~\cite{xu2024aguvis,wu2024atlas,gou2024navigating,cheng2024seeclick,gou2024navigating,uitars_1}
mainly use supervised fine-tuning on massive annotated GUI datasets, achieving strong one-step benchmark accuracy but suffering from out-of-distribution generalization and limited ability to improve through interaction with the environment.
The second wave of research, motivated by the success of DeepSeek-R1~\cite{guo2025deepseek}, has shifted toward reinforcement learning methodologies. Recent representative works~\cite{uis1,lu2025ui,luo2025gui,liu2025infigui} have adopted Group Relative Policy Optimization (GRPO)~\cite{shao2024deepseekmath}, achieving notable improvements in task completion rates.
Yet these methods treat each action as an isolated optimization target, failing to preserve sequential dependencies crucial for multi-step task execution.

\noindent
\textbf{Multi-Turn Reinforcement Learning.}
To address the limitations of single-step optimization, recent research has explored multi-turn reinforcement learning through online environment interaction~\cite{feng2025group, wang2025ragen, dong2025agentic, zhang2025landscape}. 
Recent approaches address multi-turn optimization through trajectory-aware curriculum learning~\cite{shi2025mobilegui}, stabilized data flywheels~\cite{uitars_2}, or Semi-online RL~(SO-RL) that simulates online dynamics~\cite{uis1}, though balancing deployment costs remains a challenge.
Although these methods make progress in multi-turn optimization, they still suffer from severe context inflation.

\noindent
\textbf{Vision Compression in Multimodal LLMs.}
GUI agents suffer from computational bottlenecks due to high-resolution visual histories.
Prior works address this issue via learnable query compression~\cite{hu2024bliva,li2023blip,li2024llama,zhang2025falcon,zhao2025heterogeneous} or token pruning strategies like VoCo-LLaMA~\cite{ye2025voco}.
However, these general-purpose methods often rely on static metrics or multi-stage training. In contrast, our CCPO couples visual compression with policy optimization, progressively focusing on the key regions to balance efficiency and grounding accuracy.


%% file: sec/method.tex
\subsection{Policy Optimization on Coordinates}

In GUI agent tasks, the model needs to handle multimodal multi-turn interactions and output precise action coordinates. To better align training with task success, we move beyond standard supervised fine-tuning, in which next token prediction provides a weak signal for coordinate accuracy, as well as offline RL, which fails to consider trajectory level advantages, and traditional online RL, which is inefficient and constrained by limited data. Instead, following \cite{uis1}, we use Semi-Online RL (SO-RL) by simulating online rollouts to train more efficiently while expanding the diversity of interactions.

Following the format from \cite{cheng2024seeclick, chen2025less}, we represent interaction history using A (action-only history) and AO (action with observation history) in our work. Specifically, nAO provides the agent with the most recent n observation frames together with the corresponding actions taken up to time $t$ (e.g., 4AO includes the last four screenshots and their associated actions). Previous studies \cite{uitars_1,lu2025guiodyssey, chen2025less} show that AO is more important than A because observations explicitly reveal the locality of prior actions and reduce state ambiguity.


Previous Multi-turn GUI optimization is limited by the cost of high-resolution visual tokens. Keeping full screenshot histories is inefficient, but relying only on text action traces degrades grounding and accuracy \cite{cheng2024seeclick, chen2025less, lu2025guiodyssey}. Prior work  \cite{chen2025less, lin2024showui, zhang2025ui} shows that selecting a small, and informative subset of past screenshots captures key UI changes and outperforms action-only methods, making efficient visual-history compression and selection the central challenge.

Beyond the memory and computation required to maintain a long visual history, accurate coordinate prediction is another major challenge. Because most GUI actions are coordinate-based (>\,70\%) shown in Figure~\ref{action-distribution}, accurate point prediction is essential for correct steps, reliable grounding, and overall task success. Small localization errors can cause misclicks, unintended UI changes, and disrupt the interaction flow.

Therefore, we focus on GUI grounding while reducing long-horizon costs by concentrating computation on high-confidence ROI through coordinate sampling. As shown in Figure~\ref{framework}, our approach Coordinate Compression Policy Optimization (CCPO) has three core components: Progressive Rollout Trajectory, Coordinate-Aware Spatial Compression, and Distance-Based Advantage.



\subsection{Progressive Rollout Trajectory}

We consider a GUI environment in which an agent is given a high-level instruction
\(I\) and interacts with the interface over a horizon of \(T\) steps. At each step \(t \in \{1, \dots, T\}\), the agent observes a screenshot \(s_t\) and
executes an action prediction \(a_t\), while \(a_t^\star\) represents a corresponding annotated action.
The interaction history up to step \(t\) is defined as
\begin{equation}
    h_t = \{(s_1, a_1), (s_2, a_2), \dots, (s_{t-1}, a_{t-1})\}
\end{equation}
An annotated trajectory is given by
\begin{equation}
    \tau^\star 
    = \{(s_1, a_1^\star), (s_2, a_2^\star), \dots, (s_T, a_T^\star)\}
\end{equation}
During training, we sample \(N\) rollouts from the current policy \(\pi\) where \(i = 1,\dots,N\) denotes the $i$-th rollout trajectory. 
Each action from rollout is associated with coordinates or other auxiliary information, denoted by \(c_t^{(i)}\), we define the
coordinate-augmented history as
\begin{equation}
    ch_t
    =
    \bigl\{
      (\tilde{s}_k, \{a_k^{(i)}, c_k^{(i)}\}_{i=1}^N)
    \bigr\}_{k=1}^{t-1}
\end{equation}
where \(c_t^{(i)}\) can be \(\varnothing\), depending on the action type, and \(\tilde{s}_t\) denotes the screenshot after the processing from coordinates \(c_{t-1}^{0}, \dots, c_{t-1}^{i}\). $ch_{t}$ is the $t$-th history shared across N rollouts for different $\tau_t$. \\
For a given turn \(t\), the entire rollout trajectory, for \(i = 1,\dots,N\), is
\begin{equation}
    \tau_t 
    =   \left\{ ch_t,\; \{(s_t^{1}, a_t^{1}), (s_t^{2}, a_t^{2}), \dots, (s_t^{i}, a_t^{i})\} \right\}
\end{equation}
At step \(t\) in rollout \(i\), the policy $\pi$ produces the next action conditioned on the coordinate-augmented history:
\begin{equation}
    a_t^{(i)} \sim \pi\bigl(\,\cdot \,\bigm|\, I, s_t^{(i)}, ch_t\bigr).
\end{equation}
In our multi-turn RL environment, if the annotated trajectory contains a coordinate-related action, we proceed as follows during training: we sample $N$ rollouts, each consisting of $N$ actions. Whenever an action prediction in any rollout produces coordinates, we record those coordinates as historical coordinates. After each rollout, to construct a robust attention boundary that covers potential ROI, we aggregate all coordinates from that rollout together with annotated coordinate from training data. In the next generation round, we crop the input images based on the aggregated historical coordinates collected from the previous sampling round.

The Progressive Rollout strategy offers two key benefits: $(i)$ it enables cross-rollout learning, since all rollouts share the same coordinate history, improving consistency and exploration. $(ii)$ it progressively refines this history, as each turn adds context around the annotated and predicted coordinates. This yields a confident ROI over the image derived from accumulated sampling.

\subsection{Coordinate-Aware Spatial Compression}
To enable the preservation of interaction histories within a limited context window, our CASC keeps the key visual information associated with coordinate-related actions, discarding all other historical images to achieve maximum compression.

Each action has a type and auxiliary details:
\begin{equation}
    a_t^{(i)} = \bigl(u_t^{(i)}, z_t^{(i)}\bigr),
\end{equation}
where $u_t^{(i)} \in \mathcal{U}$ is the action type (e.g., click, type, wait),
and $z_t^{(i)} \in \mathcal{Z}$ are auxiliary details (coordinates, text, time, etc.).

For general coordinate compression, we categorize actions into the following groups:\\
\textbf{Coordinate-related actions} $\mathcal{A}_{wc}$: actions such as \texttt{click}, \texttt{long-press}, \texttt{select} and \texttt{scroll}. These actions carry essential coordinate information for localization and are useful for our CASC. We treat \texttt{scroll} as a coordinate action in our work.\\
\textbf{Non-coordinate actions} $\mathcal{A}_{nc}$: actions such as \texttt{type}, \texttt{wait}, \texttt{open}, and \texttt{complete}. These actions do not require coordinate prediction, and we remove their corresponding images from the trajectory.








CASC consists of three main components that work together across multi-turn rollouts.\\
1. \textbf{Action Tracking}: This component records coordinate-based actions across turns and rollouts by tracking prior model outputs, annotated actions, and coordinate bounding boxes if present. It maintains a trajectory coordinate history for efficient reuse in later rounds.
\begin{equation}
    \mathcal{C}_t
    = \left\{\, c_t^{(i)} \;\middle|\;
        a_t^{(i)} \in \mathcal{A}_{wc};~t = 1,\dots,T
      \right\}
\end{equation}
2. \textbf{Coordinate Aggregation}:
Once coordinates have been collected, the aggregation component groups them by rollouts. It then converts the set of coordinate candidates into a single aggregated bounding box that defines a useful ROI for image history. 
\begin{equation}
    \mathcal{C}_t^{\text{anot}}
    = \left\{\, c_t^\star \;\middle|\;
        a_t^\star \in \mathcal{A}_{wc}; t = 1,\dots,T\;
      \right\}
\end{equation}
The aggregated historical coordinate set, shared across rollouts for the next round, is then
\begin{equation}
    \mathcal{C}_t^{\text{hist}}
    = {c}_t^{\star}
      \;\cup\;
      \{{c}_t^1,\dots,{c}_t^{(i)}\}
\end{equation}
3. \textbf{Region Cropping}:
For each historical step, the ROI bounding boxes are used to crop the corresponding image regions. 
These cropped regions then act as compressed image history for subsequent rounds of generation.

\begin{equation}
    \tilde{S}_t 
    = \operatorname{Crop}\big(S_t;\,\mathcal{C}_t^{\text{hist}}\big)
\end{equation}
where \(\operatorname{Crop}(\cdot;\mathcal{C}_t^{\text{hist}})\) is an operator that crops the screenshot
based on the aggregated historical coordinates \(\mathcal{C}_t^{\text{hist}}\).


By dynamically filtering out irrelevant visual redundancy while preserving task-critical context, CASC establishes a virtuous cycle where focused visual history progressively refines coordinate prediction accuracy. This approach significantly alleviates context inflation, enabling the efficient processing of interactions while maintaining precise spatial grounding.

\subsection{Distance-Based Advantage}
To provide fine-grained supervision that guides the policy toward precise locations, we design a step-level distance-based advantage. Specifically, in order to improve the grounding abilities, we use 
\begin{equation}
    r_t
    = \alpha \cdot r_{\text{format}}
    + \beta \cdot r_{\text{type}}
    + \gamma \cdot r_{\text{acc}}
\end{equation}
\textbf{Format Reward} Use $r_{\text{format}}$ to denote a binary reward: $r_{\text{format}}$ is 1 if the response follows the required format (e.g., <action> tag) and 0 otherwise.\\
\textbf{Action Type Reward} Use $r_{\text{type}}$ to denote a binary reward: assign 1 if $\hat{u} = u^*$, and 0 otherwise. \\
\textbf{Coordinate-Aware Reward (CR)}
Let $\hat{\mathbf{c}}$ be the predicted normalized coordinate, 
$\mathbf{c}$ the ground-truth normalized coordinate, and ${B}$ the
ground-truth bounding box. The normalized distance is
\begin{equation}
    d_{norm}(\hat{\mathbf{c}}, \mathbf{c}) =\left\lVert \hat{\mathbf{c}} - \mathbf{c} \right\rVert_2
\end{equation}
where $\hat{\mathbf{c}}, \mathbf{c} \in [0,1]$ and $d_{norm} \in [0,\sqrt{2}]$.
Let $\tau_{\min}$ be the normalized tolerance threshold
$\tau_{\max}$ the normalized maximum tolerance, and $w_{\min}$ the minimum weight that given only when $r_{\text{format}}$ and $r_{\text{type}}$ are correct. The coordinate accuracy reward is
\begin{equation*}
\resizebox{\columnwidth}{!}{$
r_{\text{acc}}(\hat{\mathbf{c}}, \mathbf{c}, \mathcal{B}) =
\begin{cases}
1, & d_{norm} \le \tau_{\min} \\[6pt]
w_{\min}, & d_{norm} \ge \tau_{\max} \\[6pt]
1 - \dfrac{d_{norm} - \tau_{\min}}{\tau_{\max} - \tau_{\min}}\,(1 - w_{\min}), & \tau_{\min} < d_{norm} < \tau_{\max}
\end{cases}
$}
\end{equation*}\\
The coordinate reward function is applied only when $a_{\text{type}} \in \mathcal{A}_{wc}$. For non-coordinate actions, the accuracy reward is binary:
\begin{equation}
r_{\text{acc}}(\hat a, a^*) =
\begin{cases}
1, & \text{if } (\hat u, \hat z) = (u^*, z^*), \\[4pt]
0, & \text{otherwise.}
\end{cases}
\end{equation}

Applying the coordinate-dependent advantage at the step level has two main benefits:
$(i)$ It improves performance by providing a smoother and more informative training reward instead of a hard binary reward. This is especially important given that more than half of the predictions correspond to actions with coordinates.
$(ii)$ It encourages the model to predict spatially coherent coordinates concentrated on target-relevant regions, rather than dispersed points, enabling more effective compression of long visual histories.

%% file: sec/experiment.tex
\subsection{Datasets}
We train and evaluate our models on four widely used datasets for GUI agent tasks, described below.  
\begin{table}[H]
\centering
\scalebox{0.92}{
\begin{tabular}{l l r r}
\toprule
\textbf{Dataset} & \textbf{Domain} & \textbf{Task} &  \textbf{Steps} \\
\midrule
AITW                       & Mobile \& Web  & 2,939  & 8.1  \\
Mind2Web                   & Web            & 2,350  & 7.3  \\
GUI-Odyssey                & Mobile         & 7,735  & 15.4 \\
AndroidControl             & Mobile         & 15,283 & 5.5  \\
\bottomrule
\end{tabular}}
\caption{Dataset statistics for the AITW \cite{rawles2023androidinthewild}, Mind2Web \cite{deng2023mind2webgeneralistagentweb}, GUI-Odyssey \cite{lu2025guiodyssey}, and AndroidControl \cite{li2024effects} datasets, including domain, number of tasks, and average task length (in steps).}\label{datasets}
\end{table}

\begin{table*}[ht]
\centering
\scalebox{0.8}{
\begin{tabular}{lccccccc}
\toprule
\multirow{2}{*}{\textbf{Model}} & \textbf{History Format} & \multicolumn{3}{c}{\textbf{Android Control High}} & \multicolumn{3}{c}{\textbf{GUI Odyssey}} \\
&\textit{AOT} & TM & GR & SR & TM & GR  & SR \\
\midrule

\multicolumn{7}{l}{\textit{Open-source Models}} \\
\midrule
OS-Atlas-4B ZS \cite{wu2024atlas}             & \textit{A} & 49.0 & 49.5 & 22.8 & 49.6 & 34.6 & 20.3 \\
OS-Atlas-4B FT \cite{wu2024atlas}             & \textit{A} & 84.7 & 73.8 & 67.5 & 83.5 & 61.4 & 56.4 \\
Qwen2.5VL-3B \cite{bai2025qwen2}              & \textit{A}  & 47.8 & 46.5 & 38.9 & 37.4 & 26.5 & 26.7 \\
UI-R1-3B \cite{lu2025ui}                      & -- & 57.9 & 55.7 & 45.4 & 52.2 & 34.5 & 32.5 \\
GUI-R1-3B \cite{luo2025gui}                   & \textit{A} & 58.0 & 56.2 & 46.6 & 54.8 & 41.5 & 41.3 \\
OS-Genesis-7B \cite{sun2025genesis}           & \textit{AO} & 65.9 & \textemdash & 44.4 & 11.7 & \textemdash & 3.6 \\
Aguvis-7B \cite{xu2024aguvis}                 & \textit{A} & 65.6 & \textemdash & 54.2 & 26.7 & \textemdash & 13.5 \\
GUI-R1-7B \cite{luo2025gui}                   & \textit{A} & 71.6 & 65.6 & 51.7 & 65.5 & 43.6 & 38.8 \\
AgentCPM-GUI-8B \cite{zhang2025agentcpm}      & \textit{A} & 77.7 & \textemdash & 69.2 & 90.8 & \textemdash & 75.0 \\
OS-Atlas-7B ZS \cite{wu2024atlas}             & \textit{A} & 57.4 & 54.9 & 29.8 & 60.4 & 39.7 & 27.0 \\
OS-Atlas-7B FT \cite{wu2024atlas}             & \textit{A} & 85.2 & 78.5 & 71.2 & 84.5 & 67.8 & 62.0 \\
UI-TARS-7B \cite{uitars_1}                    & \textit{AOT} & 83.7 & \textbf{80.5} & \underline{72.5} & \textbf{94.6} & \textbf{90.1} & \textbf{87.0} \\
UI-S1-7B \cite{uis1}                          & \textit{AOT} & 79.9 & 73.4 & 68.2 & 76.3 & 61.7 & 59.5 \\
\midrule
\multicolumn{7}{l}{\textit{Our Models}} \\ 
\midrule

\textbf{Qwen2.5VL-3B} (0-shot)                 & \textit{AO} & 24.9 & 68.3 & 20.2 & 27.8 & 46.4 & 14.7 \\
\quad w/ SFT                                    & \textit{AO} & 85.2 & 73.5 & 68.6 & 88.0 & 84.3 & 75.9 \\
\quad w/ Semi-online RL                         & \textit{AO} & 83.7 & 74.8 & 67.5 & 82.6 & 81.3 & 71.3 \\
\textbf{CCPO-3B-1AO}                 & \textit{AO} & 85.3 & 76.7 & 70.6 & 91.7 & 87.2 & 81.1 \\
\textbf{CCPO-3B-3AO}                 & \textit{AO} & 85.7 & 77.5 & 70.8 & 90.6 & 88.5 & 80.9 \\
\midrule
\textbf{Qwen2.5VL-7B} (0-shot)                  & \textit{AO} & 58.9 & 70.3 & 44.4 & 55.8 & 50.8 & 31.8 \\
\quad w/ SFT                                    & \textit{AO} & 85.9 & 75.9 & 70.6 & 88.0 & 84.6 & 76.0 \\
\quad w/ Semi-online RL                         & \textit{AO} & 86.3 & 76.7 & 70.6 & 89.2 & 84.9 & 76.7 \\
\textbf{CCPO-7B-1AO}                     & \textit{AO} &  \underline{86.5} & 78.8 & 72.2 & 91.1 & 87.2 & 80.3 \\
\textbf{CCPO-7B-3AO}                     & \textit{AO} & \textbf{86.9} & \underline{79.7} & \textbf{73.3} & \underline{91.8} & \underline{89.3} & \underline{82.4} \\

\bottomrule
\end{tabular}}
\caption{Results of CCPO on the Android Control and GUI-Odyssey. We report type matching (TM), grounding rate (GR), and success rate (SR). For the history format, AOT denotes \textbf{A}ction, \textbf{O}bservation, and \textbf{T}hought histories, respectively.} \label{results}
\end{table*}

\subsection{Experiments Setup}
Our experiments focus on three settings: $(i)$ SFT $(ii)$  Semi-online RL from~\cite{uis1} $(iii)$ SFT followed by CCPO.
We do not preprocess the resolution of the original screenshots. Instead, we follow the \cite{uis1} and use the same maximum pixel budget. We use a history length of 3AO as the default experimental setting unless otherwise specified. For the SFT part, we report the best checkpoint and configuration to ensure fairness. More details can be found in the Appendix~\ref{configuration}.
\subsection{Baseline Models}
Following prior work, we use Qwen2.5-VL 3B and Qwen2.5-VL 7B as our base models for both SFT and reinforcement learning. 
We compare our approach against a broad range of existing methods from two perspectives: 
$(i)$ General-purpose GUI agents that are commonly used in prior studies, like \cite{xu2024aguvis, sun2025genesis, wu2024atlas, uitars_1, uis1}.
$(ii)$ Specialized models designed to improve GUI agent efficiency, such as \cite{cheng2024seeclick, ge2025iris, lin2024showui, chen2025less}.


%% file: sec/results.tex
\begin{table*}[ht]
\centering
\setlength{\tabcolsep}{6pt}
\renewcommand{\arraystretch}{1.2}
\newcommand{\best}[1]{\textbf{#1}}
\newcommand{\secondbest}[1]{\fbox{#1}}
\scalebox{0.88}{
\begin{tabular}{lcccccc }
\toprule
\multirow{2}{*}{\textbf{Method}} & \multirow{2}{*}{\textbf{Param}}&\multicolumn{3}{c}{\textbf{Mind2Web}} &\multicolumn{2}{c}{\textbf{AITW}} \\
\cmidrule(lr){3-5}\cmidrule(lr){6-7}
&
&
\textit{Cross-Task} & 
\textit{Cross-Website} &
\textit{Cross-Domain} &
\textit{Overall} &
\textit{ClickAvg} \\
\midrule
\textbf{Qwen-VL 9.6B} \cite{bai2023qwen}            & 9.6B  & 13.3 & 9.2 & 12.0 & 54.3 & 57.4 \\
\textbf{SeeClick} \cite{cheng2024seeclick}          & 9.6B & 25.5 & 16.4 & 20.8 & 59.3 & 66.4 \\
\textbf{R-VLM} \cite{park2025r}                     & 9.6B & 28.7 & 26.1 & 24.3 & 64.9 & 71.0 \\       
\textbf{Iris} \cite{ge2025iris}                     & 9.6B & 32.0 & 26.2 & 28.8 & 63.6 & 71.0 \\
\textbf{Qwen2-VL} \cite{bai2025qwen2}               & 2B & 46.7 & 42.2 & 44.6 & 57.7 & -- \\
\textbf{ShowUI-2B} \cite{lin2024showui}             & 2B & 37.2 & 35.1 & 35.2 & 70.0 & -- \\
\textbf{SimpAgent} \cite{chen2025less}              & 2B & 48.7 & 42.2 & 45.0 & 71.5 & -- \\
\textbf{TongUI-3B} \cite{zhang2025tongui}           & 3B & 48.8 & 48.1 & 49.5 & 71.6 & -- \\
\textbf{TongUI-7B} \cite{zhang2025tongui}           & 7B & 53.4 & 49.0 & 52.9 & 73.3 & -- \\
\midrule
\textbf{Qwen2.5-VL-3B} w/ SFT                       & 3B & 52.0 & 46.5 & 48.7 & 70.8 & 78.4\\
\textbf{CCPO-3B 1AO}                                & 3B & 54.6 & 50.7 & 50.6 & 71.8 & 79.7\\
\textbf{CCPO-3B 3AO}                                & 3B & 56.5 & 51.0 & 51.8 & 73.1 & 80.4\\ \hline
\textbf{Qwen2.5-VL-7B} w/ SFT                       & 7B & 55.6 & 51.3 & 52.0 & 72.3 & 80.2\\
\textbf{CCPO-7B-1AO}                                & 7B & \underline{58.0} & \underline{53.4} & \underline{55.7} & \underline{73.5} & \underline{81.0}\\
\textbf{CCPO-7B-3AO}                                & 7B & \textbf{59.5} & \textbf{53.7} & \textbf{56.5} & \textbf{74.4} & \textbf{81.4}\\
\bottomrule
\end{tabular}}
\caption{Results of CCPO on the Mind2Web and AITW benchmarks across different settings.} \label{results2}
\end{table*}

\subsection{GUI Benchmark}

\noindent
\textbf{Android Control}
Table \ref{results} shows that adding 3AO history with CCPO delivers state-of-the-art results on Android Control (AC) for our 7B model. Specifically, it provides 3.2\% TM and 0.8\% SR improvement over UI-TARS-7B. Compared to the 1AO variant, it improves TM by 0.4\%, GR by 0.9\%, and SR by 1.1\%. It also outperforms SFT and Semi-online RL by increasing 3–4\% GR and 2–3\% SR, demonstrating consistent gains over prior work and our baselines on the AC dataset.

\noindent
\textbf{GUI Odyssey}
Table \ref{results} shows that our 3B model achieves substantial improvements over UI-S1 with gains of 15.5\%, 27.6\%, 22.9\% in TM, GR and SR with 3AO history on GUI Odyssey dataset. 
Compared to the 1AO variant, CCPO outperforms by 14.8\% in TM, 25.5\% in GR, and 20.8\% in SR. 
These results confirm that CCPO generalizes well to cross-app navigation scenarios with longer average trajectories.

\noindent
\textbf{Mind2Web} Table~\ref{results2} reports Mind2Web (M2W) results, where CCPO consistently outperforms SFT and prior baselines at both 3B and 7B models. CCPO-7B-3AO surpasses TongUI-7B by 6.1\% on Cross-Task, 4.7\% on Cross-Website, and 3.6\% on Cross-Domain, while CCPO-3B exceeds SimpAgent by 7.8\%, 8.8\%, and 6.8\%, respectively. Full results are provided in Appendix~\ref{mind2web_full}.



\noindent
\textbf{AITW} Table~\ref{results2} summarizes Android in the Wild (AITW) results, where CCPO shows consistent gains. CCPO-7B-3AO improves over CCPO-7B-1AO by 0.9\% and exceeds TongUI by 1.1\%, while CCPO-3B-3AO outperforms CCPO-3B-1AO by 1.3\%. Detailed results are in Appendix~\ref{AITW_full}.


\subsection{AO Length Scaling}
\begin{figure}[!b]
  \includegraphics[width=\linewidth]{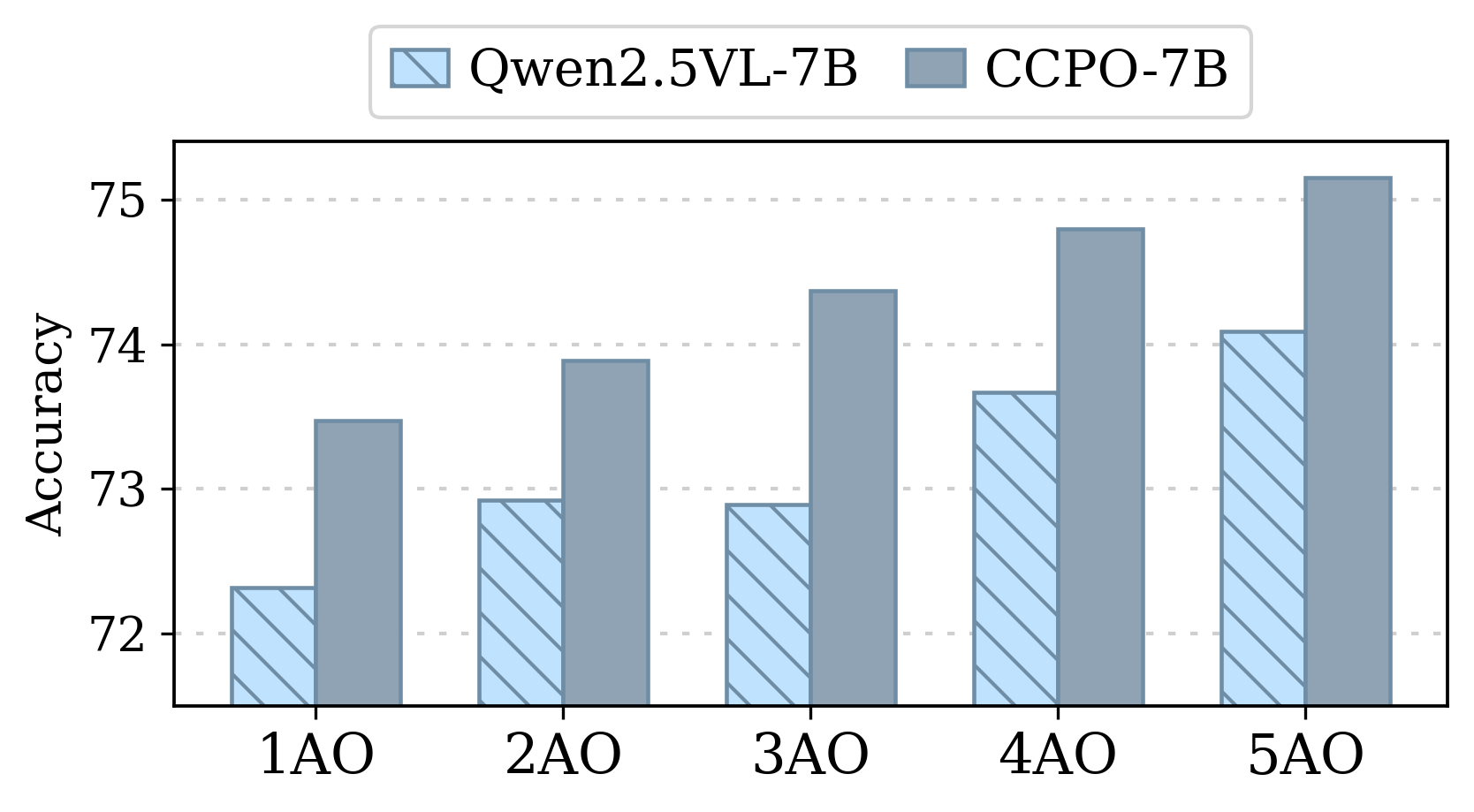} 
  \caption {Performance comparison for different AO on AITW dataset.}\label{fig:AO}
\end{figure}

Figure~\ref{fig:AO} shows how performance changes on the AITW dataset as the AO length increases from 1AO to 5AO.
For the SFT baseline, extending the AO length from 1AO to 5AO improves  accuracy by an average of 1.8\% across five subtasks. A similar pattern is observed for CCPO that 3AO consistently outperforms 1AO and 2AO, and overall accuracy improves by 1.7\%. Overall, longer AO history leads to better results for both SFT and CCPO. 
Detailed results are reported in Table~\ref{aitw-ao}.





\begin{table}[]
\centering
\small
\scalebox{0.81}{%
\begin{tabular}{lcccc}
\toprule
\textbf{Model} & \makecell{\textbf{History}\\\textbf{Length}} & \makecell{\textbf{Token}\\\textbf{Length}~$\downarrow$} & \makecell{\textbf{Compression}\\\textbf{Rate}~$\uparrow$} & \makecell{\textbf{Training}\\\textbf{Time (s/step)~$\downarrow$}} \\
\midrule
\multirow{2}{*}{SO-RL-3B} & 1AO & 6998  & 0.0\% & 515 \\
                            & 3AO & 9888 & 0.0\% & 660 \\
\midrule
\multirow{2}{*}{CCPO-3B}  & 1AO & 4271  & 38.9\%   & 154 {\scriptsize \textcolor{teal}{(3.3$\times$)}} \\
                            & 3AO & 4460  & \textbf{54.9\%}   & 174  \textbf{{\scriptsize \textcolor{teal}{(3.8$\times$)}}} \\
\midrule
\midrule 
\multirow{2}{*}{SO-RL-7B} & 1AO & 7026   & 0.0\% & 569  \\
                            & 3AO & 9550  & 0.0\% & 717 \\
\midrule
\multirow{2}{*}{CCPO-7B}  & 1AO & 4262  & 39.3\%  & 186  {\scriptsize \textcolor{teal}{(3.1$\times$)}} \\
                            & 3AO & 4473  & \textbf{53.2\%}  & 204 \textbf{{\scriptsize \textcolor{teal}{(3.5$\times$)}}} \\

\bottomrule

\end{tabular}}
\caption{
Training efficiency comparison between CCPO and Semi-online RL on the Android Control dataset.
}
\label{tab:training_efficiency}
\end{table}

\subsection{Training Efficiency}
Table~\ref{tab:training_efficiency} presents the training efficiency comparison between our method and Semi-online RL on Android Control dataset. 
Specifically, CCPO-7B compresses the token length to 39.3\%–53.2\% of the original in terms of 1AO and 3AO,
corresponding to training speedups of 3.1$\times$ and 3.5$\times$ under 1AO and 3AO settings, respectively.
Moreover, CCPO-3B achieves even greater speedups of 3.3$\times$ and 3.8$\times$ while maintaining comparable token efficiency, making it suitable for resource-constrained scenarios.
Notably, as the history length increases from 1AO to 3AO, the token length in Semi-online RL grows by 41\%, while our method maintains a relatively stable token length with only a 4\% increase. 
This demonstrates that our compression strategy scales efficiently with longer action-observation histories.
More details are available in Appendix~\ref{sec:apdx_eff}

%% file: sec/analysis.tex
\begin{table}[tp]
    \centering
    \scalebox{0.75}{
    \begin{tabular}{lccc}
        \toprule
        \textbf{Method}  & \textbf{AC-TM} & \textbf{AC-GR}  & \textbf{AC-SR} \\
        \midrule
        Qwen2.5VL-7B  SFT        & 85.94 & 75.95 & 70.60 \\
        \quad + Semi-online & 86.27 {\scriptsize \textcolor{teal}{(+0.33)}} & 77.93 {\scriptsize \textcolor{teal}{(+1.98)}} & 72.35 {\scriptsize \textcolor{teal}{(+1.75)}} \\
        \quad + CASC        & 86.72 {\scriptsize \textcolor{teal}{(+0.78)}} & 79.12 {\scriptsize \textcolor{teal}{(+3.17)}} & 72.70~~{\scriptsize \textcolor{teal}{(+2.1)}}\\
        \quad + CASC + CR   & 86.89 {\scriptsize \textcolor{teal}{(+0.95)}} & 79.71 {\scriptsize \textcolor{teal}{(+3.76)}} & 73.25 {\scriptsize \textcolor{teal}{(+2.65)}} \\
        \bottomrule
    \end{tabular}}
    \caption{Ablation study of different components on the Android Control dataset.}
    \label{tab:ablation_study}
\end{table}

\subsection{Component Ablation}
Table~\ref{tab:ablation_study} validates the contribution of each CCPO module on the Android Control dataset.
It can be observed that CASC leads to a substantial improvement in Grounding Rate, indicating that suppressing visual redundancy mitigates spatial ambiguity and enhances attention toward task-relevant regions.
The Coordinate-Aware Reward further enhances overall performance by leveraging fine-grained, distance-based supervision to achieve superior coordinate precision.

\subsection{Computational Overhead Analysis}
Table~\ref{tab:computational_overhead} shows a fine-grained profiling of the computational overhead.
Specifically, a 44\% reduction in compute load means it requires significantly fewer computational resources.
To quantify the wall-clock speedup, we measure \textit{token latency} (average time per token) and \textit{step latency} (time per training step).
In these metrics, we observe a 10\% reduction in token latency and a 35\% decrease in step latency.
These results confirm that CCPO effectively reduces training time while maintaining comparable performance.



\begin{table}[hp]
    \centering
    \scalebox{0.75}{
    \begin{tabular}{lccc}
        \toprule
        \textbf{Method} & \textbf{Compute Load} & \textbf{Token Latency} & \textbf{Step Latency} \\
         & (TFLOPS) $\downarrow$ & (ms) $\downarrow$ & (s) $\downarrow$ \\
        \midrule
        SO-RL & 9.6 & 0.064 & 297.1 \\
        \textbf{CCPO} & 5.4 {\scriptsize \textcolor{teal}{(-44\%)}} & \textbf{0.057} {\scriptsize \textcolor{teal}{(-10\%)}} & \textbf{194.5} {\scriptsize \textcolor{teal}{(-35\%)}} \\
        \bottomrule
    \end{tabular}}
    \caption{Training efficiency comparison in terms of compute load and latency.}
    \label{tab:computational_overhead}
\end{table}

\subsection{Qualitative Analysis}
\begin{figure}[ht]
    \centering
    \begin{subfigure}[b]{0.49\linewidth}
        \centering
        \includegraphics[width=\linewidth]{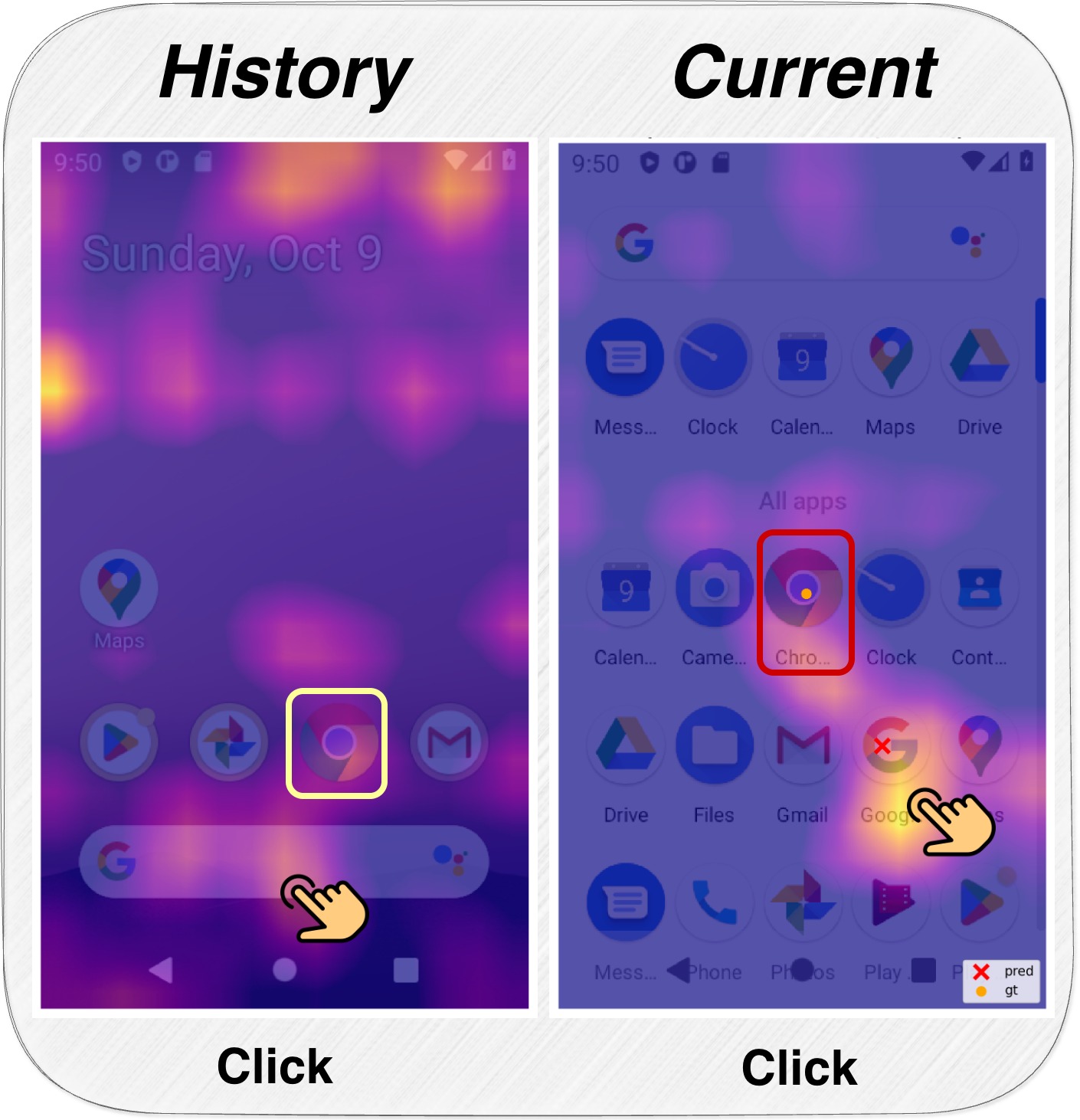} 
        \caption{SFT}
    \end{subfigure}
    \begin{subfigure}[b]{0.49\linewidth}
        \centering
        \includegraphics[width=\linewidth]{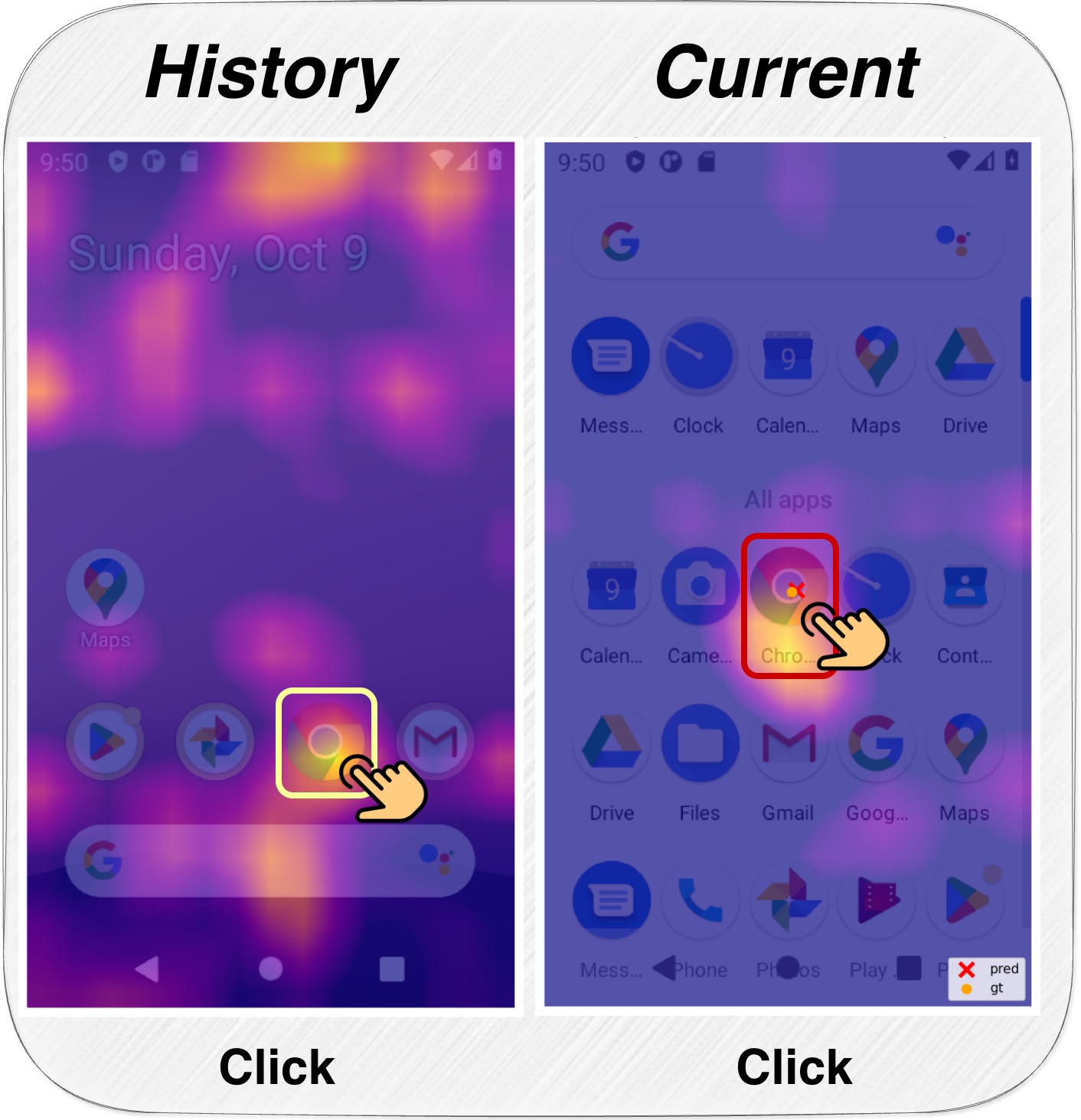} 
        \caption{CCPO}
    \end{subfigure}

        \begin{subfigure}[b]{0.49\linewidth}
        \centering
        \includegraphics[width=\linewidth]{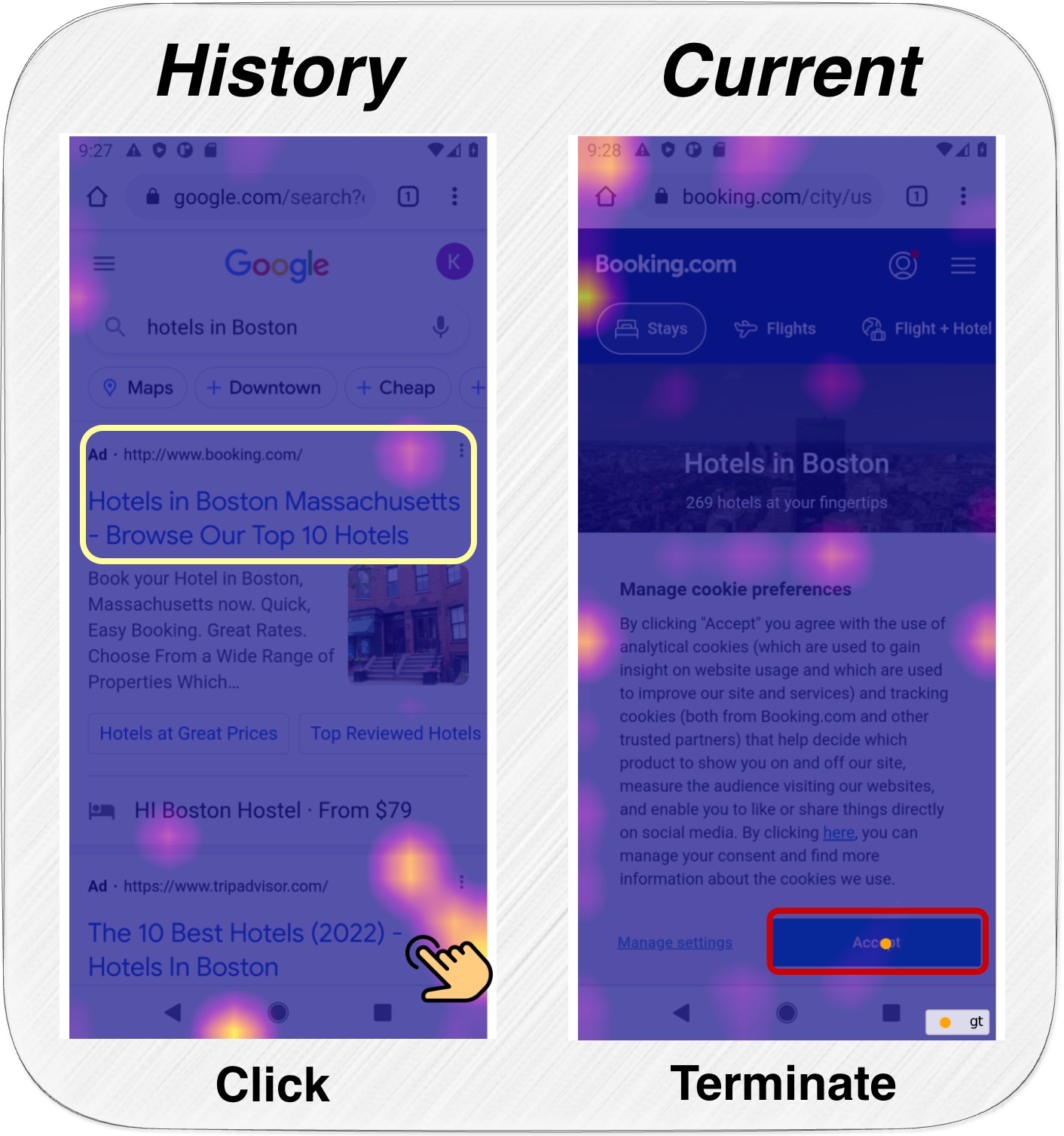} 
        \caption{SFT}
    \end{subfigure}
    \begin{subfigure}[b]{0.49\linewidth}
        \centering
        \includegraphics[width=\linewidth]{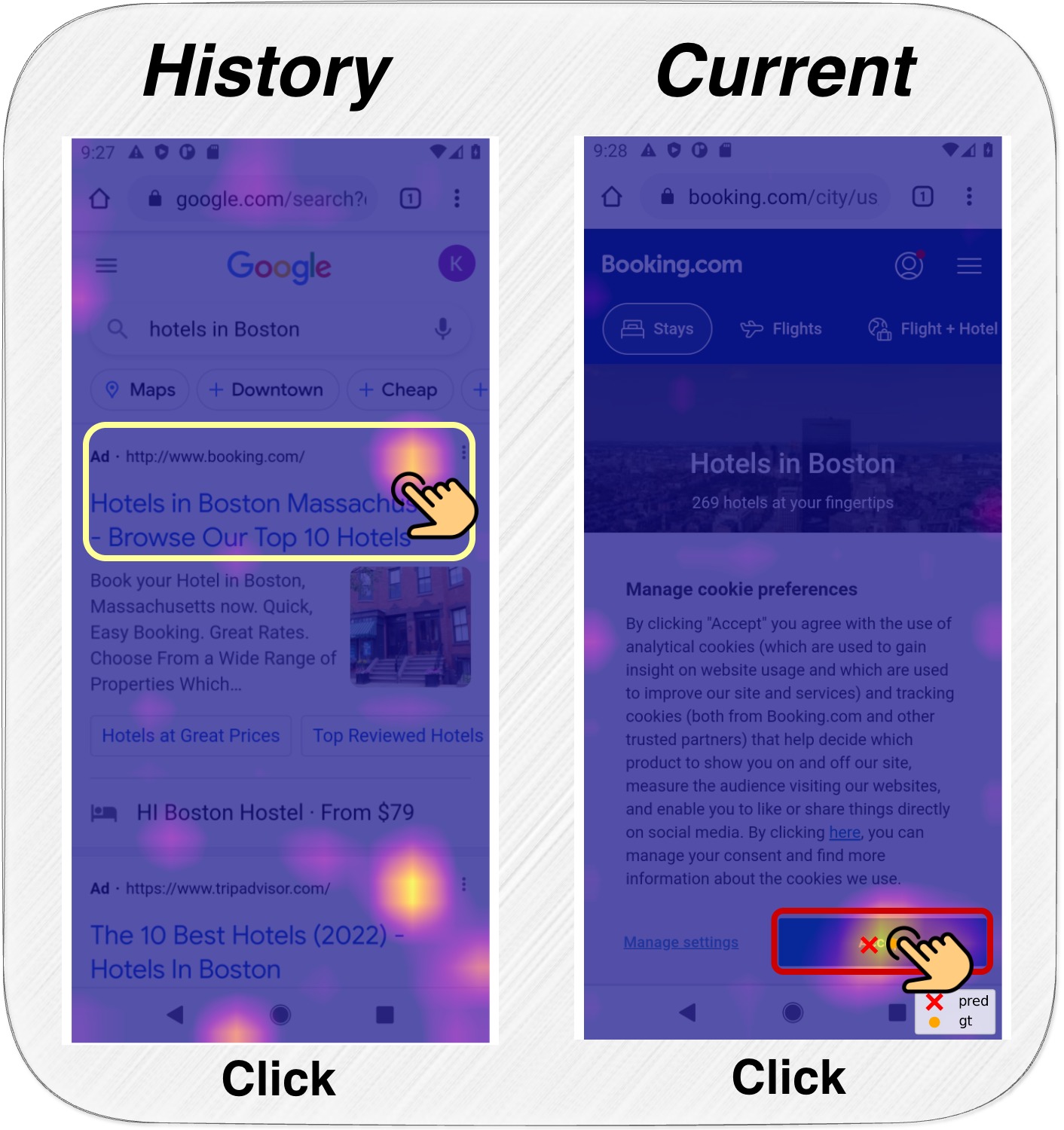} 
        \caption{CCPO}
    \end{subfigure}
    \caption{Attention maps between SFT and CCPO. CCPO accurately predicts actions and localizes coordinates, with stronger focus on detailed historical context and key elements.}
    \label{fig:attn_map}
\end{figure}
Figure~\ref{fig:attn_map} visualizes the attention maps of the SFT baseline and our CCPO method on the same GUI screenshot. The SFT model in Figure~\ref{fig:attn_map}~(a) shows scattered attention across the entire screen, while CCPO in Figure~\ref{fig:attn_map}~(b) exhibits concentrated attention on task-relevant regions. 
This demonstrates that our coordinate-based compression effectively guides the model to focus on interaction areas, thereby improving coordinate prediction accuracy and reducing computational overhead from processing irrelevant visual tokens.

%% file: sec/conclusion.tex
We introduce Coordinate Compression Policy Optimization (CCPO), an efficient policy optimization framework for GUI agents that progressively refines attention over multiple rollouts. CCPO uses Coordinate-Aware Spatial Compression (CASC) to focus on task-relevant regions from long action-observation histories. By compressing irrelevant areas, it achieves a high compression rate and substantially improves computational efficiency.
We also propose Distance-Based Advantage that guides policies smoothly toward target locations instead of relying on hard thresholds. Together, these designs enable more efficient training and stronger multi-turn decision making. CCPO achieves state-of-the-art results on diverse GUI benchmarks with fewer training resources, making it a practical and resource-efficient approach for future GUI agents.

%% file: sec/appendix.tex
\subsection{Training Configuration} \label{configuration}
Our training experiments are divided into two stages: supervised fine-tuning (SFT) and reinforcement learning (RL).

\subsubsection{Supervised Fine-tuning}
For SFT, we follow the training setup and hyperparameter configuration of SeeClick~\cite{cheng2024seeclick} and SimpAgent~\cite{chen2025less}. We freeze the visual encoder and fine-tune the model using LoRA, which enables rapid adaptation to GUI tasks. In our preliminary trials, full fine-tuning tended to overfit easily due to the short length of action output. All SFT training details are summarized in Table~\ref{SFT_config}.

\begin{table}[ht]
\centering
\scalebox{0.8}{
\begin{tabular}{l r r r r}
\toprule
\textbf{Parameter} & \textbf{AC} & \textbf{GUI O} & \textbf{AITW} & \textbf{M2W}  \\
\midrule
global\_batch\_size               & \multicolumn{1}{c}{64}    & \multicolumn{1}{c}{64} & \multicolumn{1}{c}{64} & \multicolumn{1}{c}{16} \\
total\_epochs                     & \multicolumn{1}{c}{3}     & \multicolumn{1}{c}{3} & \multicolumn{1}{c}{10} & \multicolumn{1}{c}{3}  \\
learning rate                     & \multicolumn{1}{c}{3e-5}  & \multicolumn{1}{c}{3e-4} & \multicolumn{1}{c}{3e-4} & \multicolumn{1}{c}{3e-4} \\
lr\_scheduler                      & \multicolumn{4}{c}{constant}    \\
lora rank                         & \multicolumn{4}{c}{8}    \\
lora $\alpha$                     & \multicolumn{4}{c}{16}   \\
lora module                       & \multicolumn{4}{c}{all}  \\
lora dropout                      & \multicolumn{4}{c}{0.1}  \\
warm up                           & \multicolumn{4}{c}{0.01} \\
bf16                              & \multicolumn{4}{c}{True} \\
freeze\_vision\_tower             & \multicolumn{4}{c}{True} \\
deepspeed                         & \multicolumn{4}{c}{zero2} \\
\bottomrule
\end{tabular}}
\caption{Hyperparameters for SFT Training}\label{SFT_config}
\end{table}

\subsubsection{Reinforcement Learning}
For RL training, we follow the Semi-online RL setup in~\cite{uis1}. We experiment with two settings: (1) Semi-online RL trained from scratch and (2) Continuing Semi-online RL from an SFT model. The training configurations for different datasets are summarized in Table~\ref{CCPO_config}.
Same as in~\cite{uis1}, our CCPO pipeline first performs SFT on Qwen2.5VL-3B and 7B, and then applies Semi-online RL for further optimization. 
We run our experiments on 4 nodes, each with 8× NVIDIA H200 GPUs.

\subsubsection{Data Preprocessing}
To specify our data construction in detail (e.g., using 3AO as the history representation), we generate both the training and evaluation datasets from trajectories starting from 1AO to 3AO turns, limiting the history context to at most 3AO. As a result, the training set contains instances with 1AO, 2AO, and 3AO histories, with 3AO comprising the majority. For a practical and fair comparison, the evaluation set is constructed from the same 1AO–3AO range. This preprocessing pipeline is applied uniformly across all experiments and is used to prepare data for both SFT and RL.

\begin{table}[ht]
\centering
\scalebox{0.65}{
\begin{tabular}{l r r r r}
\toprule
\textbf{Parameter} & \textbf{AC} & \textbf{GUI O} & \textbf{AITW} & \textbf{M2W} \\
\midrule
train\_batch\_size                & \multicolumn{1}{c}{32} & \multicolumn{1}{c}{8} & \multicolumn{1}{c}{16} & \multicolumn{1}{c}{16} \\
ppo\_mini\_batch\_size            & \multicolumn{1}{c}{32} & \multicolumn{1}{c}{8} & \multicolumn{1}{c}{16} & \multicolumn{1}{c}{16} \\
total\_epochs                      & \multicolumn{1}{c}{3} & \multicolumn{1}{c}{3} & \multicolumn{1}{c}{8} & \multicolumn{1}{c}{8}  \\
max\_prompt\_length               & \multicolumn{1}{c}{16384} & \multicolumn{1}{c}{32768} & \multicolumn{1}{c}{12288} & \multicolumn{1}{c}{12288} \\
DAPO threshold                    & \multicolumn{1}{c}{0.2} & \multicolumn{1}{c}{0.1} & \multicolumn{1}{c}{0.1} & \multicolumn{1}{c}{0.1}\\
reward discount (SO RL)  & \multicolumn{1}{c}{0.5} & \multicolumn{1}{c}{0.3} & \multicolumn{1}{c}{0.3} & \multicolumn{1}{c}{0.3}\\
patch threshold (SO RL)     & \multicolumn{4}{c}{1} \\
data.max\_response\_length        & \multicolumn{4}{c}{128} \\
truncation                        & \multicolumn{4}{c}{\texttt{left}} \\
use\_kl\_in\_reward               & \multicolumn{4}{c}{False} \\
Advantage weight       & \multicolumn{4}{c}{1.0} \\
historical images                 & \multicolumn{4}{c}{1 \textasciitilde ~5} \\
learning rate                     & \multicolumn{4}{c}{$5 \times 10^{-7}$} \\
fixed\_num\_mini\_batches         & \multicolumn{4}{c}{4} \\
ppo\_micro\_batch\_size\_per\_gpu  & \multicolumn{4}{c}{1} \\
kl\_loss\_coef                     & \multicolumn{4}{c}{$1 \times 10^{-4}$} \\
n\_gpus\_per\_node                 & \multicolumn{4}{c}{8} \\
nnodes                             & \multicolumn{4}{c}{4} \\


\bottomrule
\end{tabular}}
\caption{Hyperparameters for Policy Optimization Training}\label{CCPO_config}
\end{table}

\begin{figure*}[ht]
    \centering
    \begin{subfigure}[b]{0.34\textwidth}
        \centering
        \includegraphics[width=\textwidth]{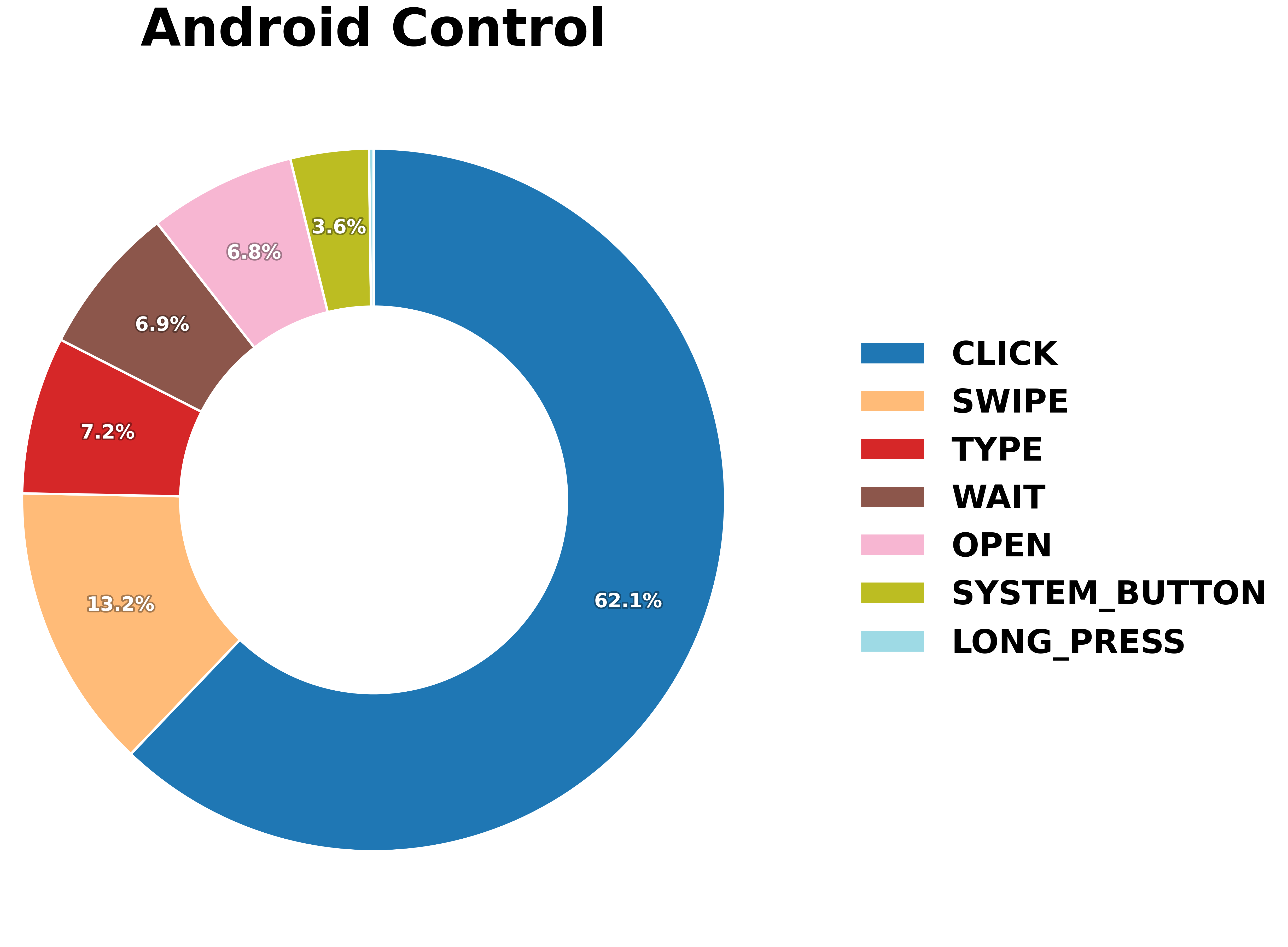} 
    \end{subfigure}
    \begin{subfigure}[b]{0.32\textwidth}
        \centering
        \includegraphics[width=\textwidth]{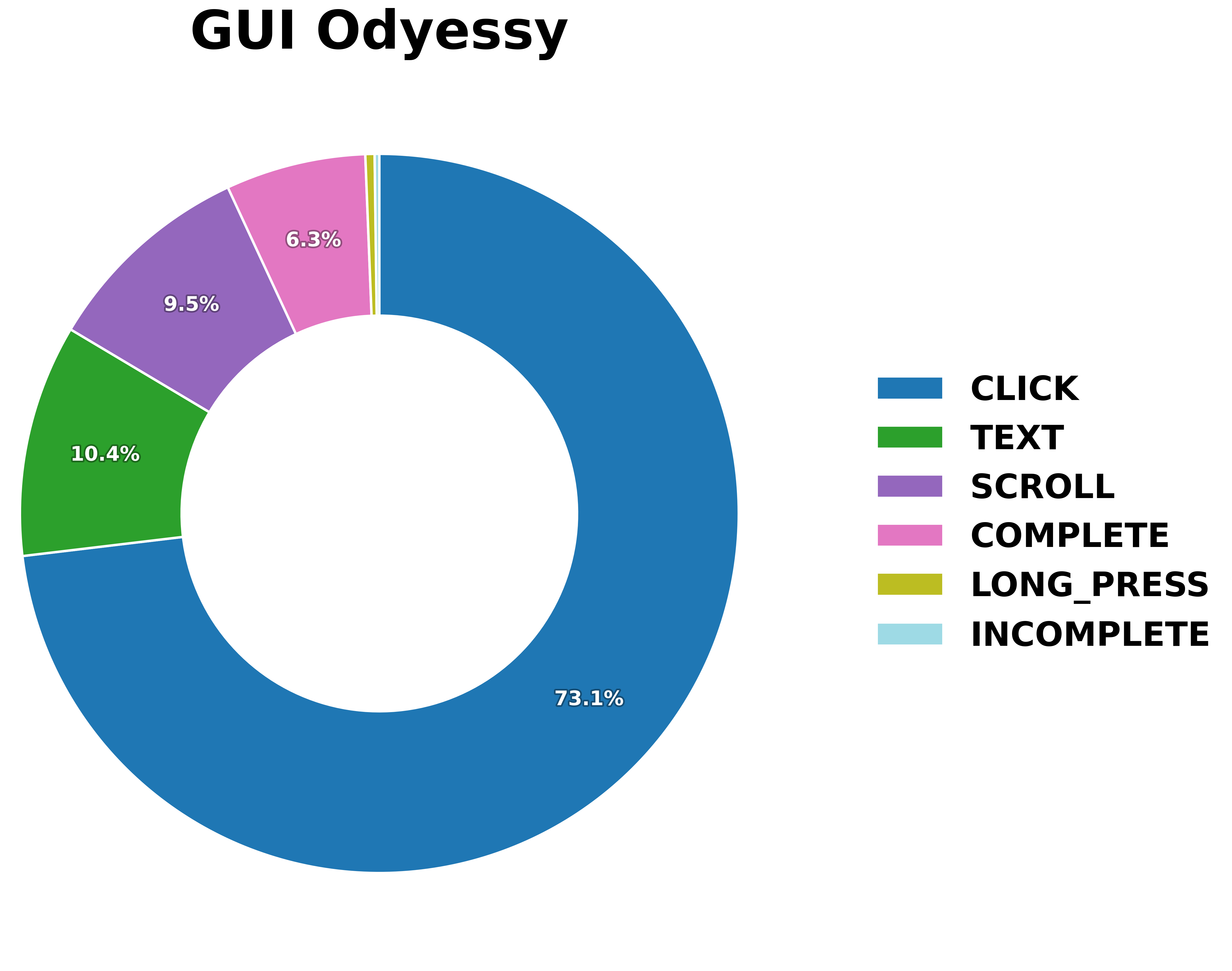} 
    \end{subfigure}
    \begin{subfigure}[b]{0.32\textwidth}
        \centering
        \includegraphics[width=\textwidth]{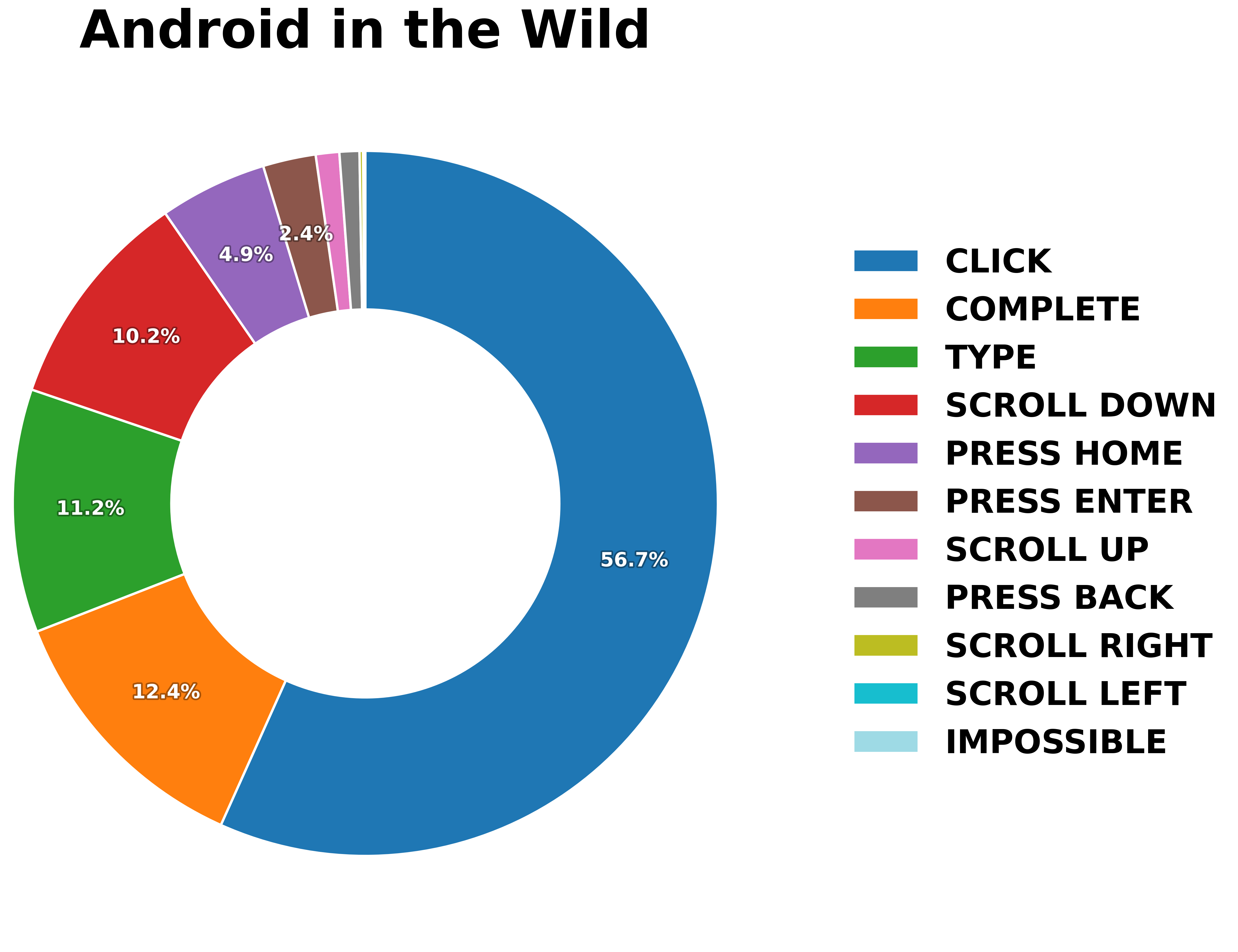} 
    \end{subfigure}
    \caption{Actions Distribution for Android Control, GUI Odyssey and Android in the Wild dataset}
    \label{action-distribution}
\end{figure*}

\subsection{Agent Tasks}

\textbf{Android Control}~\cite{li2024effects} contains 15,283 episodes covering 14,548 unique tasks across 833 Android apps, with an average of 5.5 step actions per episode. We use the official split, 13,604 episodes for training and a test set of 2,855 episodes.\\
$A_{wc}$:~\texttt{CLICK}, \texttt{LONG PRESS}, \texttt{SCROLL} \\
$A_{nc}$:~\texttt{TYPE}, \texttt{HOME}, \texttt{BACK}, \texttt{OPEN}, \texttt{WAIT} \\

\noindent\textbf{GUI Odyssey}~\cite{lu2025guiodyssey} contains 8,334 cross-app navigation episodes, with an average of 15.3 actions per episode, collected on 6 Android devices, covering 6 task categories, 212 apps, and 1,357 app combinations. \\
$A_{wc}$:~\texttt{CLICK}, \texttt{LONG PRESS}, \texttt{SCROLL} \\
$A_{nc}$:~\texttt{TYPE}, \texttt{PRESS HOME}, \texttt{PRESS BACK}, \texttt{PRESS RECENT}, \texttt{COMPLETE}, \texttt{IMPOSSIBLE} \\ 

\noindent \textbf{Android In The Wild }~\cite{rawles2023androidinthewild} is a large-scale dataset of human demonstrations for Android devices. It contains 715k interaction episodes and 30k unique instructions, spanning four multi-step subsets. We follow the dataset split of \cite{cheng2024seeclick} to evaluate models on unseen instructions and avoid overfitting as the split in previous studies.\\
$A_{wc}$:~\texttt{CLICK}, \texttt{SCROLL} \\
$A_{nc}$:~\texttt{TYPE}, \texttt{PRESS BACK}, \texttt{PRESS HOME}, \texttt{PRESS ENTER}, \texttt{STATUS TASK COMPLETE}, \texttt{STATUS TASK IMPOSSIBLE}\\

\noindent\textbf{Mind2Web}~\cite{deng2023mind2webgeneralistagentweb} is introduced as a real-world web navigation dataset aimed at training and evaluating generalist web agents. It contains more than 2000 open-ended tasks drawn from 137 real websites, where each task comes with a high-level instruction and a human demonstration trajectory. We specifically use the version from \cite{cheng2024seeclick} for a fair comparison on efficiency, rather than Multimodal Mind2Web from \cite{zheng2024seeact}. \\
$A_{wc}$: \texttt{CLICK}, \texttt{HOVER}, \texttt{ENTER}, \texttt{TYPE}, \texttt{SELECT} \\

\noindent Figure~\ref{action-distribution} summarizes the action distributions across Android Control, GUI-Odyssey, Android in the Wild, and Mind2Web. $A_{wc}$ accounts for 75\% of actions in Android Control, 83\% in GUI-Odyssey, and 67\% in AITW. In addition, Mind2Web requires coordinate prediction for every action, so $A_{wc}$ accounts for 100\%.


\subsubsection{Coordinate-Aware Reward Hyperparameters}
We analyze the Coordinate-Aware Reward hyperparameter and find that the optimal value of $\tau_{\min}$ varies across tasks.
On the AC and AITW dataset, we compare $\tau_{\min}=0.04$ and $\tau_{\min}=0.1$. Our results indicate that AITW is more challenging and benefits from a larger $\tau_{\min}$ (e.g., $\tau_{\min}=0.1$). In contrast, a smaller $\tau_{\min}$ (e.g., $\tau_{\min}=0.04$) can hinder early learning, leading to an approximate $0.2\%$ performance drop. However, for fair comparison and consistency with other datasets, we use $\tau_{\min}=0.04$ in general for all experiments reported in the main paper.

\begin{figure}[H]
  \includegraphics[width=0.9\columnwidth]{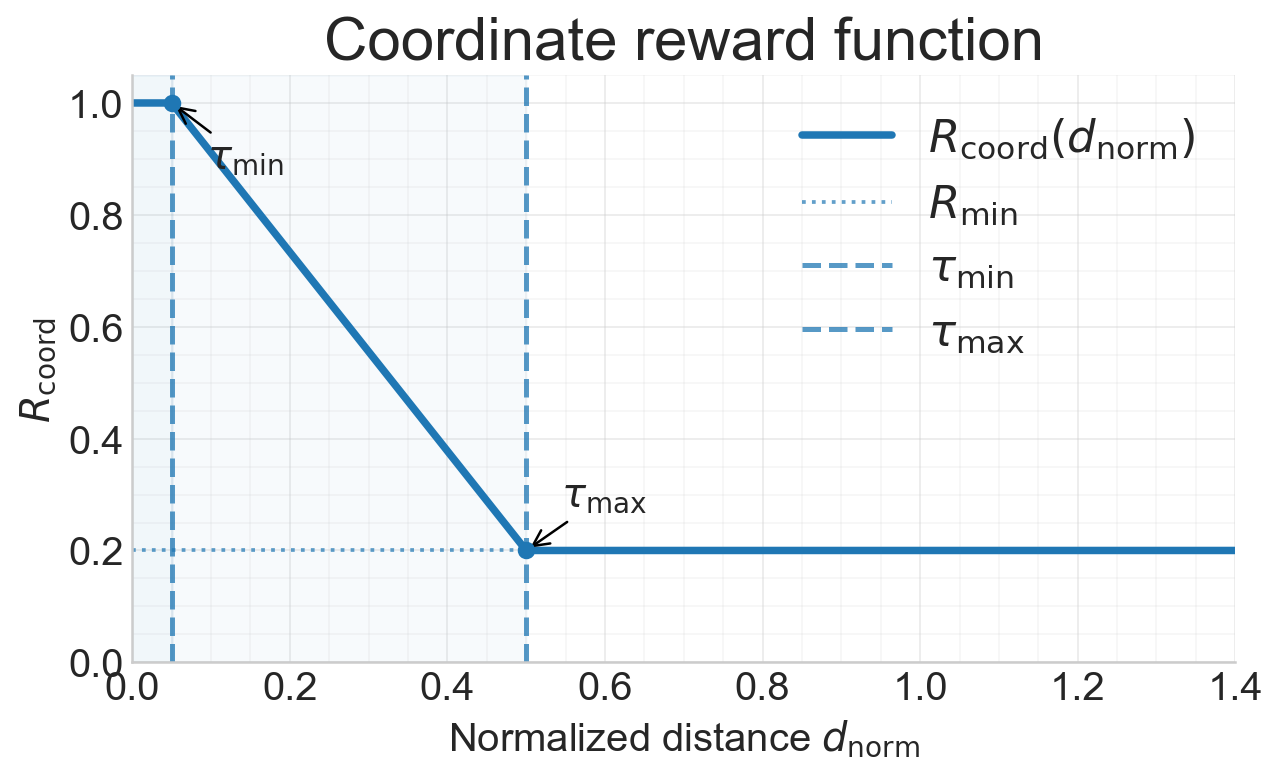} 
  \caption {Coordinate-Aware Reward Function}
\end{figure}

\newpage
\begin{figure*}[!t]
    \subsection{Ablation Study: A and AO Length Scaling}\addcontentsline{toc}{subsection}{Ablation Study: A and AO Length Scaling}
    \vspace{0.5cm}
    \centering
    \begin{subfigure}[b]{0.32\textwidth}
        \centering
        \includegraphics[width=\textwidth]{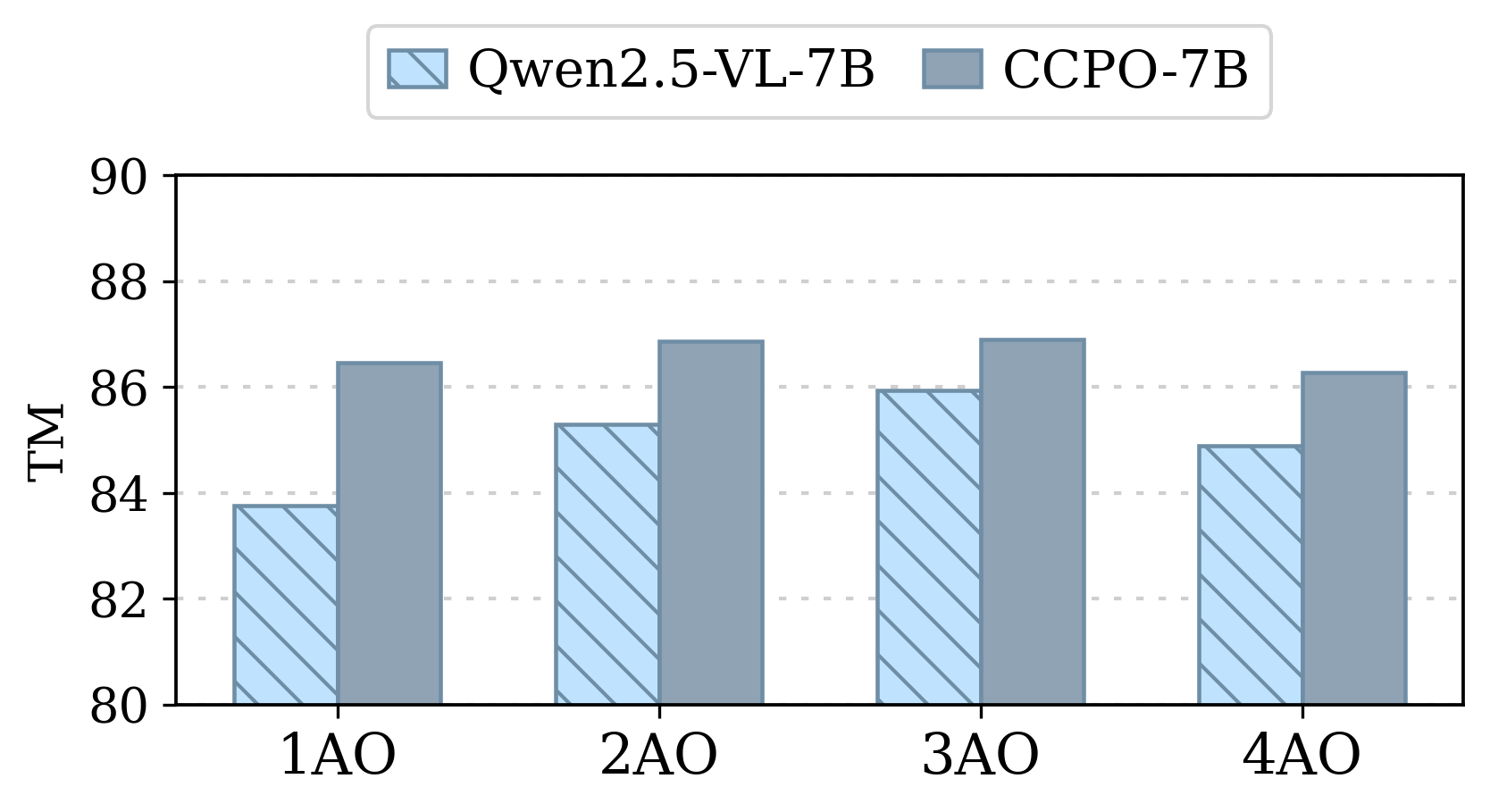} 
        \caption{Type Matching}
    \end{subfigure}
    \begin{subfigure}[b]{0.32\textwidth}
        \centering
        \includegraphics[width=\textwidth]{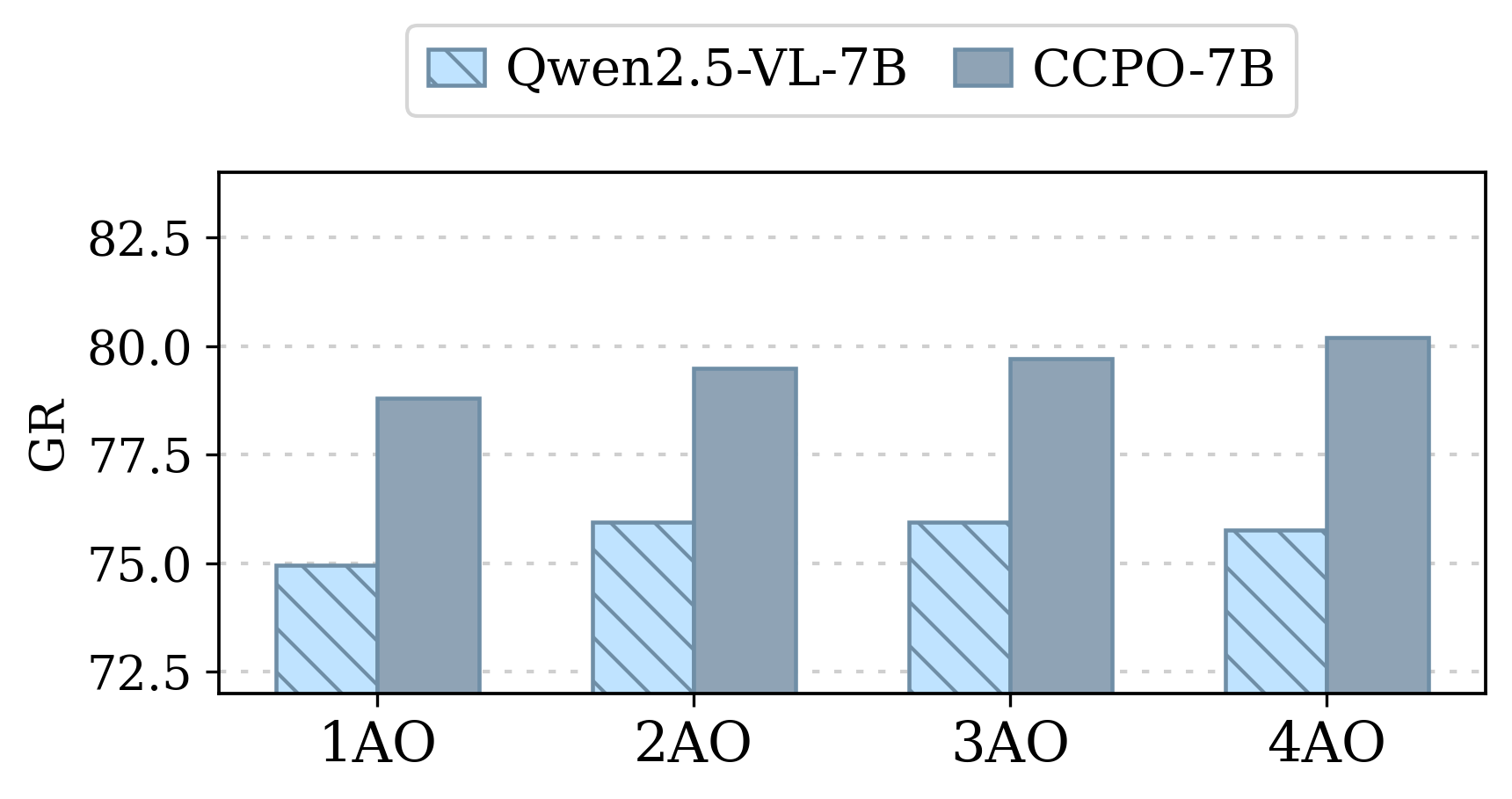} 
        \caption{Grounding Rate}
    \end{subfigure}
    \begin{subfigure}[b]{0.32\textwidth}
        \centering
        \includegraphics[width=\textwidth]{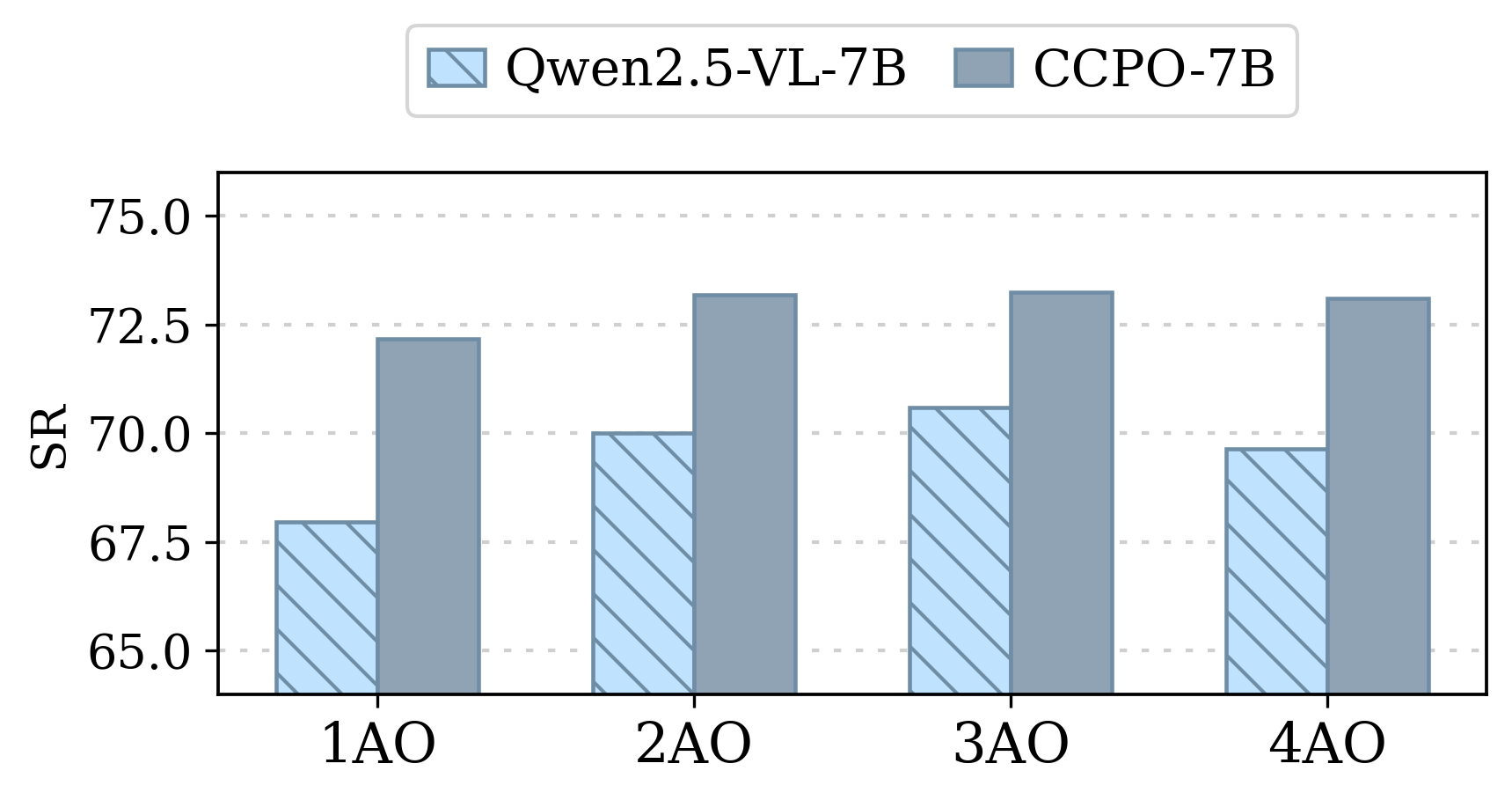} 
        \caption{Step Rate}
    \end{subfigure}
    \caption{Performance on the AC dataset across different AO.}
    \label{fig:AC_AO}
\end{figure*}
We analyze the AO setting on the AC datasets and compare different AO lengths by reporting the TM, GR, and SR results in Figure~\ref{fig:AC_AO} and Table~\ref{tab:AC_ao}. The results indicate that the optimal AO length for AC is around 3 or 4. Notably, for grounding ability, AC with 4AO outperforms 3AO, suggesting that our CCPO effectively improves grounding performance.

Table~\ref{aitw-ao} compares AITW performance across different AO lengths in detail. Our results show that AO length of 5 achieves the best performance. It achieves 1.65\% improvement over CCPO-7B-1AO across five subtasks on average.

\begin{figure*}[!b]
    \centering
    \begin{subfigure}[b]{0.45\textwidth}
        \centering
        \includegraphics[width=\textwidth]{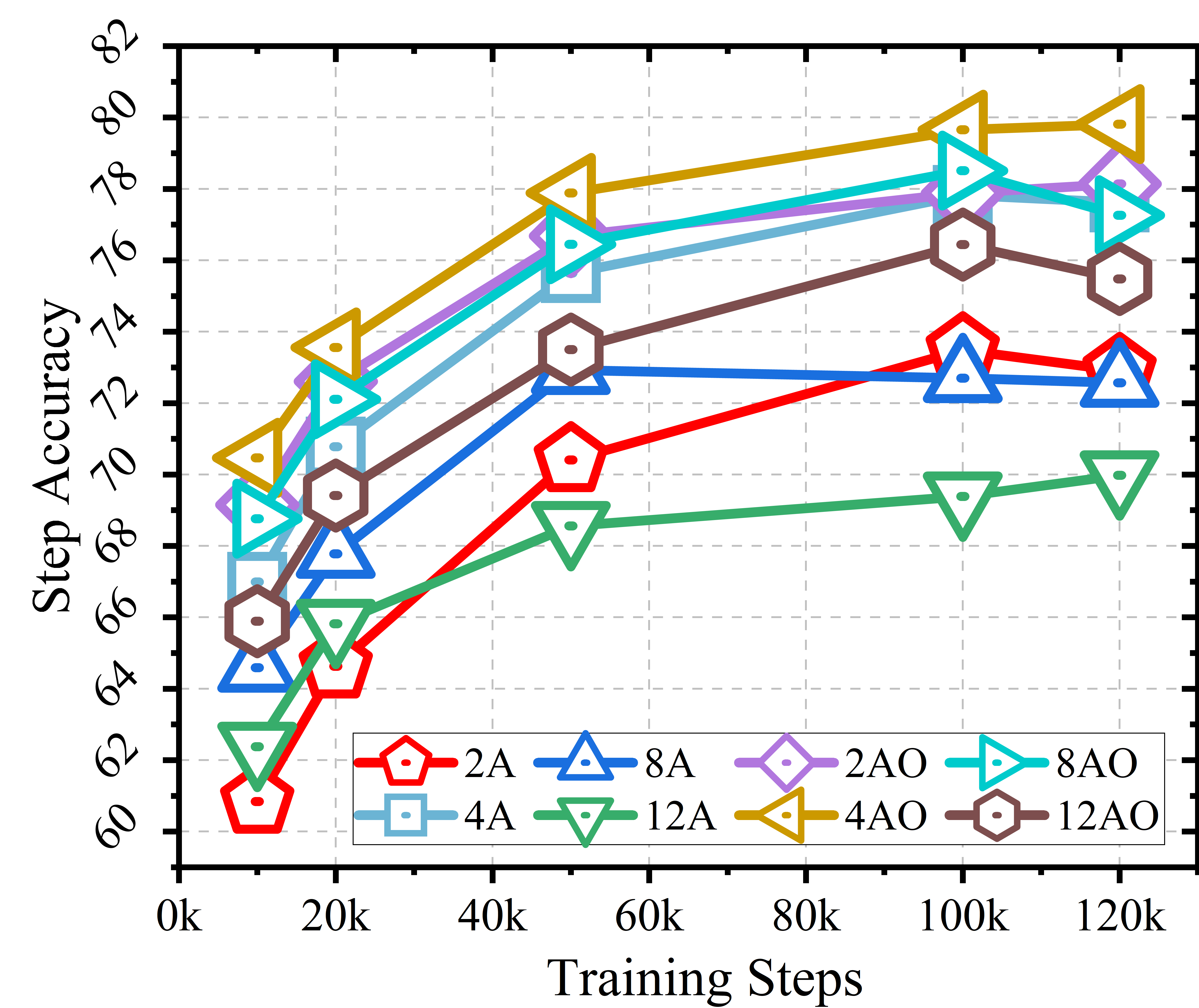} 
    \end{subfigure}
    \hspace{0.8cm}
    \begin{subfigure}[b]{0.45\textwidth}
        \centering
        \includegraphics[width=\textwidth]{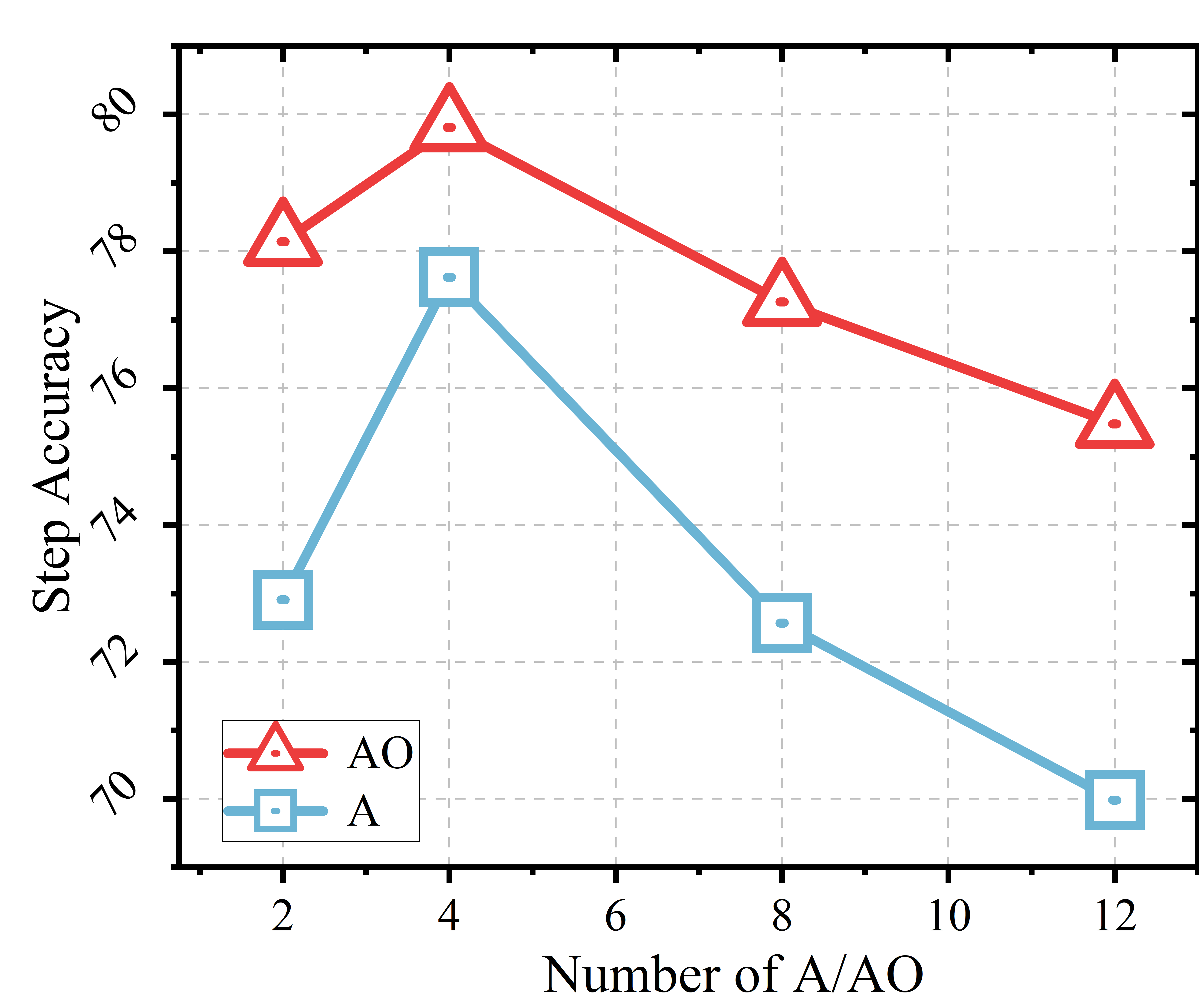} 
    \end{subfigure}
    \caption{Performance on the GUI Odyssey dataset across different A and AO in SFT training.}
    \label{fig: gui_o_AO}
\end{figure*}

Beyond the AO settings explored on the AITW and AC datasets, we also evaluate the detailed A and AO settings on GUI Odyssey without CCPO in Figure~\ref{fig: gui_o_AO}. Since GUI Odyssey involves much longer trajectories than AITW and AC, we investigate action lengths of 2, 4, 8, and 12. Overall, AO consistently outperforms A under the same length setting. However, a longer AO is not necessarily better. We observe that 2A (2AO) achieves performance comparable to 8A (8AO), while 4A (4AO) yields the best results among its tested lengths. This suggests that the optimal AO length for GUI Odyssey is around 4 and simply increasing the AO length does not necessarily yield better performance. More broadly, the best choice of A and AO length is task-dependent. However, compressing the historical images efficiently does not degrade performance.





\vspace{-4cm} 

\begin{table}[]
  \centering
\scalebox{0.92}{
  \begin{tabular}{llccc}
    \toprule
    \textbf{Model} & \textbf{AO} & \textbf{TM} & \textbf{GR} & \textbf{SR} \\
    \midrule
    \multirow{4}{*}{\textit{Qwen2.5-VL-7B}}
      & 1AO & 83.75 & 74.95 & 67.97 \\
      & 2AO & 85.30 & 75.95 & 70.00 \\
      & 3AO & 85.94 & 75.95 & 70.60 \\
      & 4AO & 84.89 & 75.77 & 69.65 \\
    \midrule
    \multirow{4}{*}{\textit{CCPO-7B}}
      & 1AO & 86.45 & 78.80 & 72.18 \\
      & 2AO & 86.86 & 79.48 & 73.19 \\
      & 3AO & \textbf{86.89} & 79.71 & \textbf{73.25} \\
      & 4AO & 86.27 & \textbf{80.20} & 73.11 \\
    \bottomrule
  \end{tabular}}
  \caption{Performance comparison on Android Control from 1AO to 4AO.} \label{tab:AC_ao}
\end{table}

\clearpage
\raggedbottom
\begin{table*}[t]
    \subsection{Ablation Study: Compression Variants}\addcontentsline{toc}{subsection}{Ablation Study: Compression Variant}
   \vspace{0.5cm}
  \centering
  \begin{tabular}{llccccc}
    \toprule   
    \textbf{Model} & \textbf{Variant} & \textit{TM} & \textit{GR} & \textit{SR} &  \textit{Compression Rate} $\uparrow$ & \textit{Step Time} $\downarrow$ \\
    \midrule
      Qwen2.5-VL-7B SO-RL-1AO & ORIG  & 84.40 & 75.86 & 68.62  & 0.0\%  & 569  \\
      CCPO-7B-1AO             & MIN  & \textbf{86.50} & 78.52 & 72.07 & 30.6\%  & 327 {\scriptsize \textcolor{teal}{(1.7$\times$)}} \\
      CCPO-7B-1AO             & MAX  & 86.45 & \textbf{78.80}  & \textbf{72.18}  & \textbf{39.3\%}  & 186 \textbf{{\scriptsize \textcolor{teal}{(3.1$\times$)}}}\\ 
      \midrule
      Qwen2.5-VL-7B SO-RL-3AO & ORIG & 86.26 & 76.72 & 70.58   & 0.0\%  & 717  \\
      CCPO-7B-3AO             & MIN  & 86.77 & 79.32 & 72.88 & 37.7\% & 410 {\scriptsize \textcolor{teal}{(1.7$\times$)}} \\
      CCPO-7B-3AO             & MAX  & \textbf{86.89} & \textbf{79.71}  & \textbf{73.25} & \textbf{53.2\%} & 204 \textbf{{\scriptsize \textcolor{teal}{(3.5$\times$)}}}\\
    \bottomrule
  \end{tabular}
  \caption{Results on the AC dataset under different compression variants} \label{min-max}
\end{table*}
To better understand how compression affects CCPO performance, we implement two variants: \textbf{MAX-COMPRESS} and \textbf{MIN-COMPRESS}. MAX-COMPRESS is the version we used in the main paper, while MIN-COMPRESS is explored as an additional study. The key difference lies in how screenshots from non-coordinate actions are handled: \\
MAX-COMPRESS: retains only visuals from coordinate-related actions ($A_{wc}$), discarding all other historical images.
MIN-COMPRESS: compresses only visuals from coordinate-related actions, leaving visuals from non-coordinate actions ($A_{nc}$) unchanged. \\
Empirically, in Table~\ref{min-max}, MIN-COMPRESS achieves performance close to MAX-COMPRESS, with worse grounding ability. However, it yields a worse compression ratio and longer training time.
These results suggest that screenshots from non-coordinate actions $A_{nc}$ contribute little to performance. In contrast, coordinate-related visual information appears essential, improving grounding and the overall success rate.

\subsection{Ablation Study: Rollout}
Since our method progressively aggregates rollouts into trajectories, the number of rollouts is a key hyperparameter that can affect performance. We evaluate rollout counts of 2, 4, 8, 12, and 16 while keeping the number of training epochs  in Table~\ref{rollout-results}. Our results show that a rollout of around 12 performs best on AITW, with 8 also performing strongly. To ensure a fair comparison with prior work~\cite{uis1}, we use 8 rollouts for our main results. Notably, larger rollout counts tend to improve performance, likely because more rollouts yield ROI regions that are more precise and estimated with higher confidence.

\vspace{-5cm}
\begin{table*}[b]
  \centering
  \scalebox{1.0}{
  \begin{tabular}{llcccccc}
    \toprule
    \textbf{Model} & \textbf{Rollouts} & \textit{General} & \textit{Single} & \textit{Web Shopping} &  \textit{Install} & \textit{Google Apps} & \textit{Overall} \\
    \midrule
    \multirow{5}{*}{\textit{CCPO-7B-3AO}}
      & 2 & 65.56  & 78.91  & 66.63 & 78.05  & 77.43  & 73.31 \\
      & 4 & 67.47 & 79.38 & 68.48 & 77.41 & 77.23 & 73.99 \\
      & 8 & \textbf{68.29} & 78.67 & 69.62 & 77.25 & \textbf{78.02} & 74.37 \\
      & 12 & 66.75 & \textbf{79.62} & \textbf{70.57} & \textbf{78.61} & 77.23 & \textbf{74.56} \\
      & 16 & 66.86  & 79.38  & 70.33 & 77.89  & 77.43  & 74.38  \\
    \bottomrule
  \end{tabular}}
  \caption{Results comparison on AITW dataset across rollout from 2 to 16.} \label{rollout-results}
\end{table*}

\clearpage

\begin{figure*}[!b]
    \centering
    \begin{subfigure}[b]{0.48\textwidth}
        \centering
        \includegraphics[width=\textwidth]{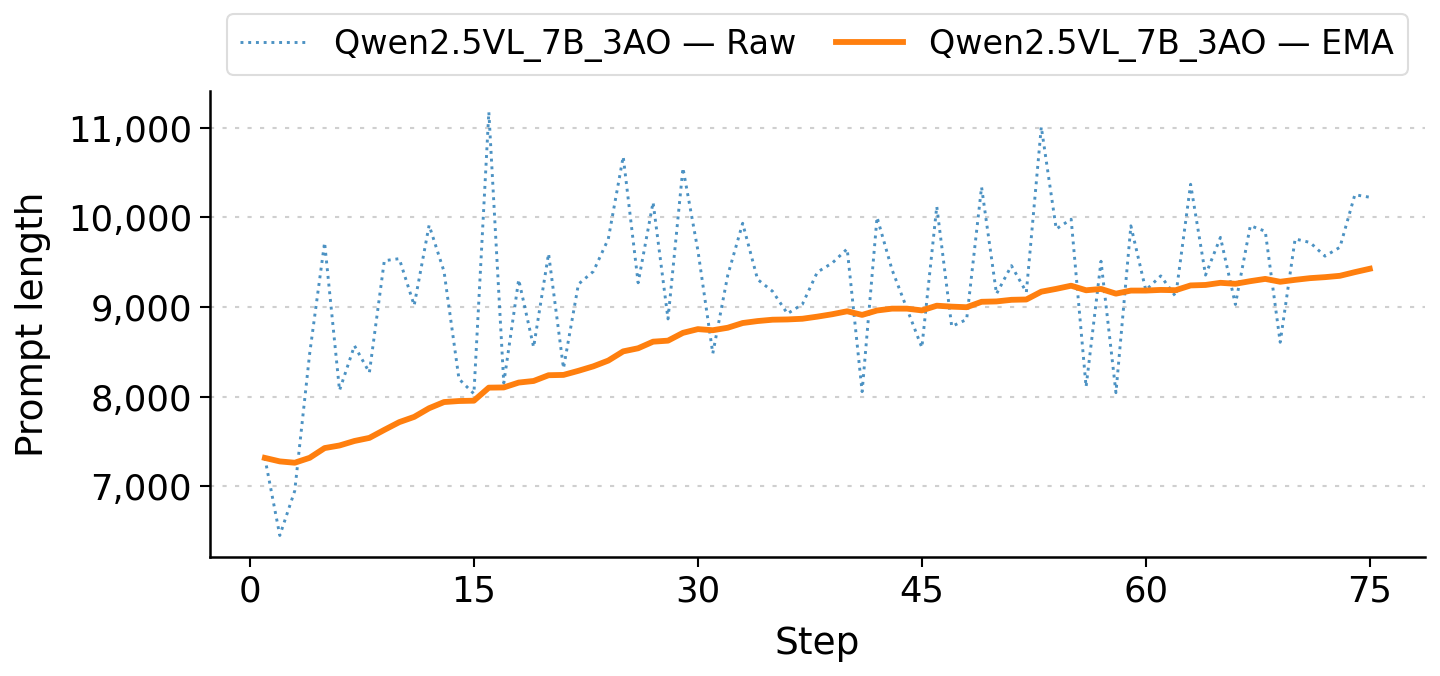} 
    \end{subfigure}
    \hspace{0.2cm}
    \begin{subfigure}[b]{0.48\textwidth}
        \centering
        \includegraphics[width=\textwidth]{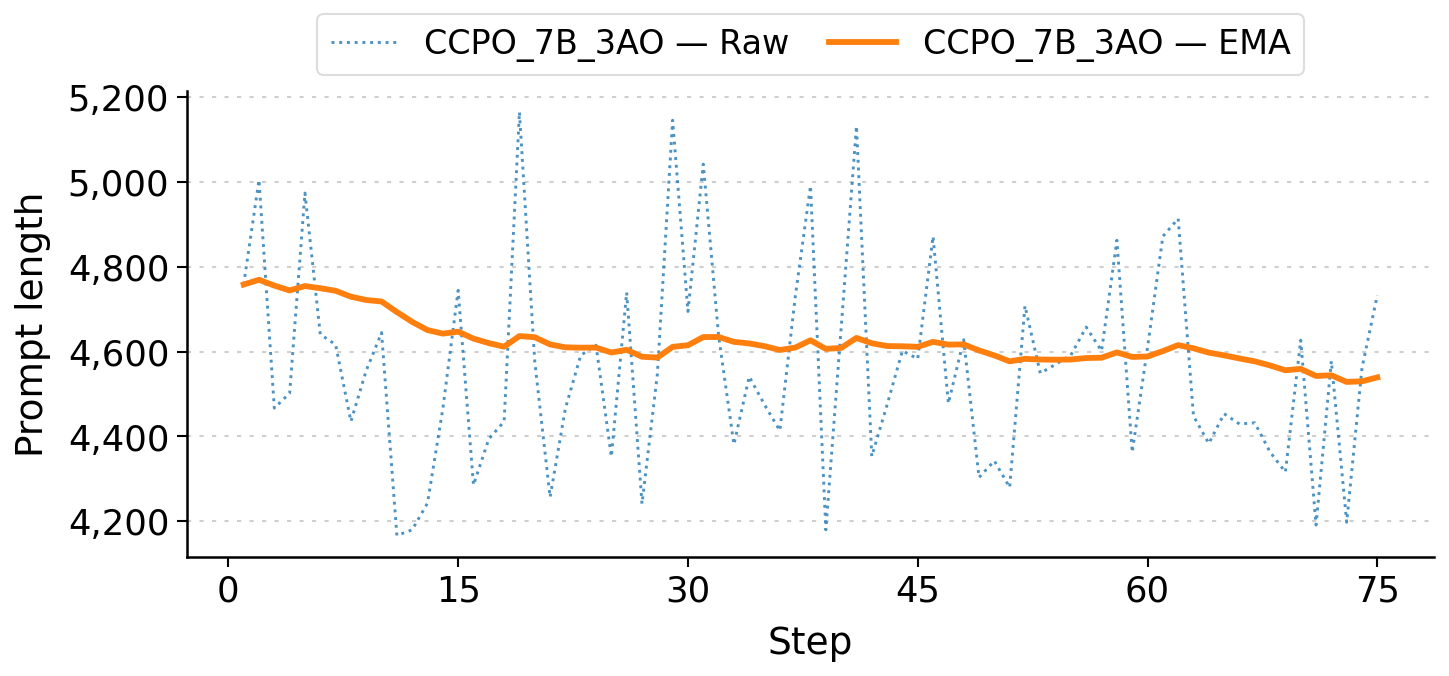} 
    \end{subfigure}
    \caption{Token length comparison during the training.}
    \label{fig: prompt_length}
\end{figure*}

\begin{table*}[t]
  \centering
      \subsection{Ablation Study: Efficiency}\addcontentsline{toc}{subsection}{Ablation Study: Efficiency}
      \vspace{0.5cm}
      \label{sec:apdx_eff}
\scalebox{1.0}{
  \begin{tabular}{llcccc}
    \toprule
    \textbf{Model} & \textbf{AO} & \textit{Overall} &  \textit{Token Length}    & \textit{Compression Rate} $\uparrow$ & \textit{Step Time} $\downarrow$ \\ 
    \midrule
    \multirow{5}{*}{\makecell{\textit{Qwen2.5VL-7B}\\\textit{SO-RL}}}
      & 1AO & 69.84 & 2180 & 0\% & 140\\
      & 2AO & 69.74 & 2613 & 0\% & 175 \\
      & 3AO & 69.98 & 2895 & 0\% & 204 \\
      & 4AO & 70.00  & 2964 & 0\% & 213 \\
      & 5AO & 70.53 & 3065 & 0\% & 225 \\
    \midrule
    \multirow{5}{*}{\textit{CCPO-7B}}
      & 1AO & 73.47 & 1463 & 32.9\% & 114 {\scriptsize \textcolor{teal}{(1.2$\times$)}} \\
      & 2AO & 73.89 & 1475 & 43.5\% & 116 {\scriptsize \textcolor{teal}{(1.5$\times$)}}  \\
      & 3AO & 74.37 & 1530 & 46.1\% & 120 {\scriptsize \textcolor{teal}{(1.7$\times$)}}  \\
      & 4AO & 74.80 & 1555 & 47.5\% & 121 \textbf{{\scriptsize \textcolor{teal}{(1.8$\times$)}}} \\
      & 5AO & \textbf{75.12} & 1577 & \textbf{49.3\%} & 128 \textbf{{\scriptsize \textcolor{teal}{(1.8$\times$)}}} \\
    \bottomrule
  \end{tabular}}
  \caption{Performance and efficiency comparison of the Qwen2.5VL-7B Semi-online Reinforcement learning and CCPO-7B model on AITW across 1AO–5AO settings.} \label{aitw-compression}
\end{table*}




We further evaluate efficiency and performance on the AITW dataset (Table~\ref{aitw-compression}). Because the AITW images are much smaller than the AC images, the input token sequences are correspondingly shorter. Despite this, the compression rate remains largely unchanged. CCPO maintains a high compression ratio of around 50\%, comparable to its compression rate on the AC in Table~\ref{ac-compression}, while still delivering a training speedup of 1.8$\times$. In general, performance improves as compression becomes more effective.

Figure~\ref{fig: prompt_length} illustrates an important trend in Semi-online reinforcement learning. As the success rate increases, the prompt length tends to grow. Trajectories are rolled out until termination or until the model makes an incorrect prediction, so more accurate agents typically produce longer successful trajectories and therefore longer inputs. In contrast, CCPO can slightly reduce the prompt length over training, benefiting from increasingly accurate predictions. More accurate coordinate predictions enable more aggressive historical image pruning, which in turn shortens the prompt length.

Finally, we report detailed compression rates and training efficiency on the AC datasets in Table~\ref{ac-compression}. CCPO achieves up to 55\% token-length reduction and 3.5 $\times$ speedup for 4AO in comparison to original Semi-online RL.



\begin{table*}[t]
  \centering
  \small
  \scalebox{0.95}{
  \begin{tabular}{lcccccccc}
    \toprule
    \textbf{Model} &
    {\textit{M2W-Task}} &
    {\textit{M2W-Domain}} &
    {\textit{M2W-Website}} &
    {\textit{AITW}} &
    {\textit{Token Length}} &
    {\textit{Compression Rate} $\uparrow$} &
    {\textit{FLOPs (T)}} \\
    \midrule
    SFT-1AO  & 55.60 & 51.97 & 51.34 & 72.32 & 2662 & 0.0\%  & 20.75 \\
    SFT-3AO  & 57.30 & 52.20 & 54.69 & 72.89 & 4293 & 0.0\%  & 33.05 \\
    CCPO-1AO* & 56.62 & \textbf{54.53} & 52.09 & \textbf{73.64} & 1634 & 38.6\% & 12.84 {\scriptsize \textcolor{teal}{(1.6$\times$)}} \\
    CCPO-3AO* & \textbf{57.36} & 55.14 & \textbf{52.51} & 73.29 & 1718 & \textbf{60.0\%} & 13.47 \textbf{{\scriptsize \textcolor{teal}{(2.5$\times$)}}} \\
    \bottomrule
  \end{tabular}}
  \caption{Performance of 7B model with optimized inference on Mind2Web and AITW datasets.} \label{optim_inference}
\end{table*}

\subsubsection{Inference Efficiency}
To evaluate the practical deployment potential, we investigate the performance of "optimized inference" (denoted by * in Table~\ref{optim_inference}). Note that while our main experiments strictly follow prior works for fair comparison, this experiment explores the efficiency limit by aligning inference with our training-time CASC. 
Specifically, Table~\ref{optim_inference} indicates that resolving the inconsistency between training and inference results in substantial efficiency improvements. CCPO-3AO achieves a 60.0\% compression rate, translating to a 2.5$\times$ reduction in FLOPs compared to SFT. 
This confirms that full screenshots contain significant visual noise, and our CASC strategy effectively acts as a denoising mechanism that purifies task-critical information. 

\begin{table*}[t]
  \centering
\scalebox{0.9}{
  \begin{tabular}{llcccccc}
    \toprule
    \textbf{Model} & \textbf{AO} & \textit{TM}  & \textit{GR}  & \textit{SR} &  \textit{Token Length}    & \textit{Compression Rate} $\uparrow$ & \textit{Step Time} $\downarrow$ \\
    \midrule
    \multirow{2}{*}{\makecell{\textit{Qwen2.5VL-3B}\\\textit{SO-RL}}}
      & 1AO & 82.16 & 74.09 & 62.94 & 6998 & 0\% & 515 \\
      & 3AO & 83.70 & 74.78 & 67.45 & 9888 & 0\% & 660 \\
    \midrule
    \multirow{2}{*}{\textit{CCPO-3B}}
      & 1AO & 85.33 &  76.72 & 70.60 & 4271 & 39.0\% & 154 {\scriptsize \textcolor{teal}{(3.3$\times$)}}\\
      & 3AO & \textbf{85.72}  & \textbf{77.49} & \textbf{70.79} & 4460 & \textbf{54.9\%} & 174  \textbf{{\scriptsize \textcolor{teal}{(3.8$\times$)}}}\\
    \bottomrule
  \end{tabular}}
  \caption{Performance and efficiency comparison of the Qwen2.5VL-3B Semi-online Reinforcement learning and CCPO-3B model on the AC 1AO and 3AO settings.} \label{3B-ao-comparsion}
\end{table*}

\begin{table*}[t]
  \centering
\scalebox{0.9}{
  \begin{tabular}{llcccccc}
    \toprule
    \textbf{Model} & \textbf{AO} & \textit{TM}  & \textit{GR}  & \textit{SR} &  \textit{Token Length}    & \textit{Compression Rate} $\uparrow$ & \textit{Step Time} $\downarrow$ \\ 
    \midrule
    \multirow{4}{*}{\makecell{\textit{Qwen2.5VL-7B}\\\textit{SO-RL}}}
      & 1AO & 84.40 & 75.86 & 68.62 & 7026 & 0\% & 569 \\
      & 2AO & 85.05 & 76.32 & 70.04 & 8482 & 0\% & 661 \\
      & 3AO & 86.26 & 76.72 & 70.58 & 9550 & 0\% & 717 \\
      & 4AO & 85.65 & 76.74 & 70.48 & 10089 & 0\% & 760 \\
    \midrule
    \multirow{4}{*}{\textit{CCPO-7B}}
      & 1AO & 86.45 & 78.80 & 72.18  & 4263 & 39.33\% & 186 {\scriptsize \textcolor{teal}{(3.1$\times$)}}\\
      & 2AO & 86.86 & 79.48 & 73.19  & 4384 & 48.31\% & 196  {\scriptsize \textcolor{teal}{(3.4$\times$)}}\\
      & 3AO & \textbf{86.89} & 79.71 & \textbf{73.25} & 4474 & 53.15\% & 204  \textbf{{\scriptsize \textcolor{teal}{(3.5$\times$)}}}\\
      & 4AO & 86.27 & \textbf{80.20} & 73.11& 4531 & \textbf{55.01\%} & 220 \textbf{{\scriptsize \textcolor{teal}{(3.5$\times$)}}}\\
    \bottomrule
  \end{tabular}}
  \caption{Performance and efficiency comparison of the Qwen2.5VL-7B Semi-online Reinforcement learning and CCPO-7B model on the AC across 1AO–4AO settings.} \label{ac-compression}
\end{table*}

\begin{table*}[t]
  \centering
\scalebox{1.0}{
  \begin{tabular}{llcccccc}
    \toprule
    \textbf{Model} & \textbf{AO} & \textit{General} & \textit{Single} & \textit{Web Shopping} &  \textit{Install} & \textit{Google Apps} & \textit{Overall} \\
    \midrule
    \multirow{5}{*}{\textit{Qwen2.5VL-7B}}
      & 1AO & 64.85 & 77.49 & 68.54 & 76.86 & 73.86 & 72.32 \\
      & 2AO & 65.80 & 75.83 & 69.74 & 77.17 & 76.04 & 72.92 \\
      & 3AO & 65.32 & 76.78 & 70.04 & 76.46 & 75.84 & 72.89 \\
      & 4AO & 66.51 & 77.25 & 70.93 & 77.41 & 76.24 & 73.67 \\
      & 5AO & 66.27 & 78.67 & 70.93 & 77.33 & 77.23 & 74.09 \\
    \midrule
    \multirow{5}{*}{\textit{CCPO-7B}}
      & 1AO & 66.98  & 78.20  & 68.66 & 77.25  & 76.24  & 73.47 \\
      & 2AO & 66.03 & 78.67 & 69.86 & 78.05 & 76.83 & 73.89 \\
      & 3AO & 68.29 & 78.67 & 69.62 & 77.25 & 78.02 & 74.37 \\
      & 4AO & \textbf{68.53} & \textbf{78.91} & 70.63 & 77.09 & 78.81 & 74.80 \\
      & 5AO & 67.58 & 78.44 & \textbf{71.35} & \textbf{78.85}  & \textbf{79.41}  & \textbf{75.12} \\
    \bottomrule
  \end{tabular}}
  \caption{Performance comparison of the Qwen2.5VL-7B SFT and CCPO-7B model on AITW across 1AO–5AO settings.} \label{aitw-ao}
\end{table*}

\clearpage


\begin{table*}[t]
\centering
\setlength{\tabcolsep}{6pt}
\renewcommand{\arraystretch}{1.2}
\newcommand{\best}[1]{\textbf{#1}}
\newcommand{\secondbest}[1]{\fbox{#1}}
\scalebox{0.88}{
\begin{tabular}{lccccccc }
\toprule
\textbf{Method} &
\textit{General} &
\textit{Single} &
\textit{Web Shopping} & 
\textit{Install} &
\textit{Google Apps} &
\textit{Overall} &
\textit{ClickAvg} \\
\midrule
\textbf{Qwen-VL 9.6B} \cite{bai2023qwen}            & 49.5 & 64.7 & 50.7 & 59.9 & 46.9 & 54.3 & 57.4 \\
\textbf{SeeClick} \cite{cheng2024seeclick}          & 54.0 & 73.7 & 57.6 & 66.4 & 54.9 & 59.3 & 66.4 \\
\textbf{R-VLM} \cite{park2025r}                     & 59.9 & 72.5 & 61.7 & 70.6 & 59.6 & 64.9 & 71.0 \\       
\textbf{Qwen2-VL} \cite{bai2025qwen2}               & 48.3 & 57.8 & 51.6 & 77.4 & 52.9 & 57.7 & -- \\
\textbf{Iris} \cite{ge2025iris}                     & 61.5 & 71.4 & 58.3 & 66.4 & 60.2 & 63.6 & 71.0 \\
\textbf{ShowUI-2B} \cite{lin2024showui}             & 63.9 & 77.5 & 66.6 & 72.5 & 69.7 & 70.0 & -- \\
\textbf{SimpAgent} \cite{chen2025less}              & 64.1 & 76.2 & 67.2 & 75.8 & 74.0 & 71.5 & -- \\
\textbf{TongUI-3B} \cite{zhang2025tongui}           & 65.6 & 77.0 & 65.8 & 75.1 & 74.5 & 71.6 & -- \\
\textbf{TongUI-7B} \cite{zhang2025tongui}           & \underline{67.6} & \textbf{79.9} & 69.1 & 76.3 & 73.5 & 73.3 & -- \\
\midrule
\textbf{Qwen2.5-VL-3B} w/ SFT                       & 61.52 & 75.35 & 67.22 & 75.81 & 74.05 & 70.79 & 78.42 \\
\textbf{CCPO-3B 1AO} w/o CR                         & 62.71 & 78.20 & 65.07 & 75.50	& 76.44 & 71.58 & 79.12 \\
\textbf{CCPO-3B 1AO}                                & 64.25 & 76.07 & 67.22 & 76.14 & 75.44 & 71.83 & 79.71 \\
\textbf{CCPO-3B 3AO} w/o CR                         & 65.20 & 79.15 & 66.63 & 76.54 & 75.84 & 72.67 & 79.99 \\
\textbf{CCPO-3B 3AO}                                & 65.32 & 77.49 & 68.30 & \textbf{78.29} & 76.04 & 73.09 & 80.42\\ \midrule
\textbf{Qwen2.5-VL-7B} w/ SFT                       & 64.84 & 77.48 & 68.54 & 76.85 & 73.86 & 72.31 & 80.24\\
\textbf{CCPO-7B 1AO} w/o CR                         & 66.39 & 79.38 & 67.46 & 75.90	& 76.24 & 73.07 & 79.34 \\
\textbf{CCPO-7B-1AO}                                & 66.98 & 78.19 & 68.66 & 77.25 & 76.24 & \underline{73.46} & \underline{80.98}\\
\textbf{CCPO-7B 3AO} w/o CR                         & 64.85 & \underline{79.38} & \textbf{69.98} & 77.25	& \textbf{79.01} & 74.09 & 80.52 \\
\textbf{CCPO-7B-3AO}                                & \textbf{68.28} & 78.67 & \underline{69.61} & \underline{77.25} & \underline{78.02} & \textbf{74.37} & \textbf{81.38}\\
\bottomrule
\end{tabular}}
\caption{Evaluation results for CCPO on the AITW benchmark.} \label{AITW_full}
\end{table*}

\begin{table*}[t]
\centering
\scalebox{0.75}{
\begin{tabular}{l c c c c c c c c c c}
\toprule
\multirow{2}{*}{\textbf{Method}} & \multirow{2}{*}{\textbf{Param.}} &
\multicolumn{3}{c}{\textbf{Cross-Task}} &
\multicolumn{3}{c}{\textbf{Cross-Website}} &
\multicolumn{3}{c}{\textbf{Cross-Domain}} \\
\cmidrule(lr){3-5}\cmidrule(lr){6-8}\cmidrule(lr){9-11}
& & Ele.Acc & Op.F1 & Step SR & Ele.Acc & Op.F1 & Step SR & Ele.Acc & Op.F1 & Step SR \\
\midrule
Qwen-VL  \cite{bai2023qwen}                   & 9.6B & 15.9 & 86.7 & 13.3 & 13.2 & 83.5 &  9.2 & 14.1 & 84.3 & 12.0 \\
CogAgent \cite{hong2024cogagent}              & 18B  & 22.4 & 53.0 & 17.6 & 18.4 & 42.4 & 13.4 & 20.6 & 42.0 & 15.5 \\
SeeClick \cite{cheng2024seeclick}             & 9.6B & 28.3 & 87.0 & 25.5 & 21.4 & 80.6 & 16.4 & 23.2 & 84.8 & 20.8 \\
R-VLM \cite{park2025r}                        & 9.6B & 31.6 & 88.0 & 28.7 & 29.5 & 84.9 & 26.1 & 26.7 & 85.3 & 24.3 \\    
Iris  \cite{ge2025iris}                       & 9.6B & 33.5 & 87.1 & 32.0 & 31.2 & 82.2 & 26.2 & 32.8 & 85.1 & 28.8 \\
Qwen2-VL \cite{bai2025qwen2}                  & 2B   & 51.6 & 88.6 & 46.7 & 48.5 & 85.7 & 42.2 & 48.3 & 87.0 & 44.6 \\
ShowUI   \cite{lin2024showui}                 & 2B   & 39.9 & 88.6 & 37.2 & 41.6 & 83.5 & 35.1 & 39.4 & 86.8 & 35.2 \\
SimpAgent  \cite{chen2025less}                & 2B   & 52.4 & 89.4 & 48.7 & 48.2 & 85.8 & 42.2 & 49.0 & 88.2 & 45.0 \\
TongUI-3B \cite{zhang2025tongui}              & 3B   & 53.4 & 89.0 & 48.8 & 54.2 & 86.4 & 48.1 & 53.8 & 88.2 & 49.5 \\
TongUI-7B \cite{zhang2025tongui}              & 7B   & 58.1 & 88.7 & 53.4 & 55.6 & 87.2 & 49.0 & 57.6 & 88.7 & 52.9 \\
\midrule
Qwen2.5-VL 3B 1AO w/SFT                       & 3B   & 56.61 & 89.98 & 52.01 & 53.31 & 86.59 & 46.49 & 53.00 &	87.98 & 48.69 \\               
CCPO-3B 1AO                                   & 3B   & 58.66 & 90.12 & 54.55 & 56.52 & 87.77 & 50.62 & 54.51 & 87.99 & 50.58 \\
\midrule  
Qwen2.5-VL 3B  3AO w/SFT                      & 3B   & 57.91 & 90.98 & 53.23 & 53.00 &86.89 & 46.52 & 53.19	& 88.75 & 49.37 \\
CCPO-3B 3AO                                   & 3B   & 61.03 & 90.95 & 56.50  & 58.41 & 86.61 & 50.97 &56.23 & 88.80 & 51.76 \\
\midrule  
Qwen2.5-VL 7B 1AO w/SFT                       & 7B   & 59.67 & 90.63 & 55.60 & 56.77 & 88.44 & 51.34 & 56.11 & 88.64 &51.97 \\
CCPO-7B 1AO                                   & 7B   & 62.14 & 91.10 & 58.00 & 59.66 & 86.89 & 53.41 & 59.71 & 90.23 & 55.66 \\
\midrule   
Qwen2.5-VL 7B 3AO w/SFT                       & 7B   & 61.92 & 91.28 & 57.30 & 59.10 & 87.51 & 52.20 & 59.03 & 90.32	 & 54.69 \\
CCPO-7B 3AO                                   & 7B   & \textbf{64.31} & \textbf{91.78} & \textbf{59.51} & \textbf{60.14} & \underline{87.78} & \underline{53.65} & \textbf{60.79} & \textbf{90.58} & \textbf{56.49} \\
\bottomrule
\end{tabular}}
\caption{Performance comparison on Mind2Web across different settings. We report element accuracy (Ele.Acc), operation F1 (Op.F1), and step success rate (Step SR). } \label{mind2web_full}
\label{tab:mind2web}
\end{table*}

\begin{algorithm*}[ht]
\caption{CCPO-Style RL Training Loop}
\label{alg:dapo-training}
\begin{algorithmic}[1]
\Require Actor policy $\pi_\theta$, critic $V_\phi$, reference policy $\pi_{\mathrm{ref}}$,
         reward function $r$, validation reward $r_{\mathrm{val}}$, training prompts $\mathcal{D}$,
         batch size $B$, epochs $E$, discount $\gamma$, GAE parameter $\lambda$, DAPO threshold $\tau$, 
         critic warmup $K$.
\State Initialization, global step $s \gets 0$
\For{$e = 1$ to $E$}
    \For{mini-batch of prompts $\mathcal{B} \subset \mathcal{D}$}
        \State $\mathcal{T} \gets \emptyset$ \Comment{accumulated trajectories}
        \State $n_{\mathrm{prompts}} \gets 0$, $n_{\mathrm{gen}} \gets 0$
        \Repeat \Comment{multi-round rollouts with \textcolor{purple}{coordinate tracking}}
            \State $\tilde{\mathcal{B}} \gets \textsc{PrepareBatch}(\mathcal{B})$
            \State \Comment{\textcolor{purple}{aggregate coordinates} from previous steps}
            \For{\textcolor{blue}{each trajectory $t$ in $\tilde{\mathcal{B}}$}}
                \State $g \gets \text{example group of } t$
                \State $\mathcal{C}_g \gets \textsc{\textcolor{blue}{AggregateCoordinates}}(g)$ \Comment{\textcolor{purple}{collect coords} from previous \textcolor{purple}{resps} and \textcolor{purple}{ref}}
                \State $\mathcal{B}_g \gets \textsc{\textcolor{blue}{CoordsToBboxes}}(\mathcal{C}_g)$ \Comment{\textcolor{purple}{convert to bboxes}}
                \State $\tilde{\mathcal{B}}[t] \gets \textsc{\textcolor{blue}{CropImages}}(\tilde{\mathcal{B}}[t], \mathcal{B}_g)$ \Comment{\textcolor{purple}{apply bbox cropping}}
            \EndFor
            \State $\hat{\mathcal{T}} \gets \textsc{Rollout}(\pi_\theta, \tilde{\mathcal{B}})$
            \State $\hat{\mathcal{T}} \gets \textsc{ActionRewards}(\hat{\mathcal{T}}, r)$ 
            \State $\hat{\mathcal{T}} \gets \textsc{\textcolor{blue}{CoordsRewards}}(\hat{\mathcal{T}}, r)$\Comment{\textcolor{purple}{compute coords weight}}
            \If{KL penalty enabled}
                \State $\hat{\mathcal{T}} \gets \textsc{ApplyKLPenalty}(\hat{\mathcal{T}}, \pi_\theta, \pi_{\mathrm{ref}})$
            \EndIf
            \State Update coordinate history from $\hat{\mathcal{T}}$
            \State $\mathcal{T} \gets \mathcal{T} \cup \hat{\mathcal{T}}$
            \State Update $n_{\mathrm{prompts}}$, $n_{\mathrm{gen}}$
        \Until{DAPO stopping condition or max generations reached}
        
        \If{DAPO enabled}
            \State $\mathcal{T} \gets \textsc{DAPOFilter}(\mathcal{T}, \tau, B)$ \Comment{keep top-$B$ by DAPO}
        \Else
            \State $\mathcal{T} \gets \textsc{TruncateToBatchSize}(\mathcal{T}, B)$ 
        \EndIf
        
        \State $\mathcal{T} \gets \textsc{ComputeMasksAndLengths}(\mathcal{T})$ \Comment{build response masks, sequence lengths}
        \State $\mathcal{T} \gets \mathcal{T} \cup \log \pi_\theta(\text{tokens} \mid \text{inputs})$ \Comment{store old log-probs for PPO-style}
        \If{reference policy used}
            \State $\mathcal{T} \gets \mathcal{T} \cup \log \pi_{\mathrm{ref}}(\text{tokens} \mid \text{inputs})$
        \EndIf
        \If{critic used}
            \State $\mathcal{T} \gets \mathcal{T} \cup V_\phi(\text{states})$
        \EndIf
        \State $\mathcal{T} \gets \textsc{\textcolor{blue}{ComputeAdvantages}}(\mathcal{T}, \gamma, \lambda, \text{estimator})$\Comment{\textcolor{purple}{compute episode level advantages}}
        \If{critic used}
            \State $\phi \gets \textsc{UpdateCritic}(\mathcal{T}, \phi)$
        \EndIf
        \If{$s \ge K$}
            \State $\theta \gets \textsc{UpdateActor}(\mathcal{T}, \theta)$
        \EndIf
        \State $s \gets s + 1$
    \EndFor
\EndFor
\State \Return $\pi_\theta, V_\phi$
\end{algorithmic}
\end{algorithm*}

\clearpage

\begin{figure*}[ht]
    \setlength{\abovecaptionskip}{0.01cm}
    \setlength{\belowcaptionskip}{0.01cm}
    \subsection{Use Case}\addcontentsline{toc}{subsection}{Use Case}
    \vspace{0.5cm}
    \centering
    \begin{subfigure}[b]{\textwidth}
        \centering
        \setlength{\abovecaptionskip}{0.01cm}
        \setlength{\belowcaptionskip}{0.01cm}
        \includegraphics[width=0.71\textwidth]{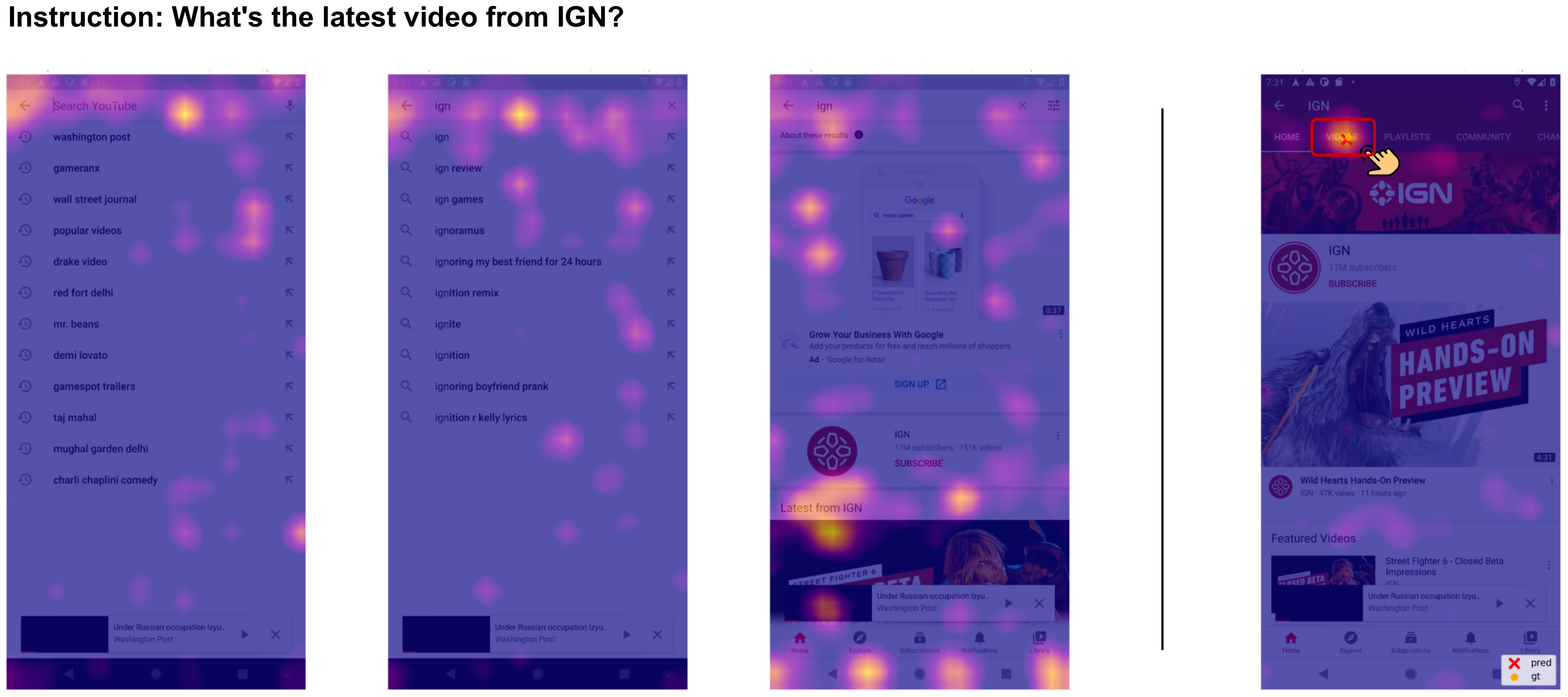} 
        \caption{CCPO 7B 3AO}
    \end{subfigure}
    \vspace{0.2cm}
    \begin{subfigure}[b]{\textwidth}
        \centering
        \setlength{\abovecaptionskip}{0.01cm}
        \setlength{\belowcaptionskip}{0.01cm}
        \includegraphics[width=0.71\textwidth]{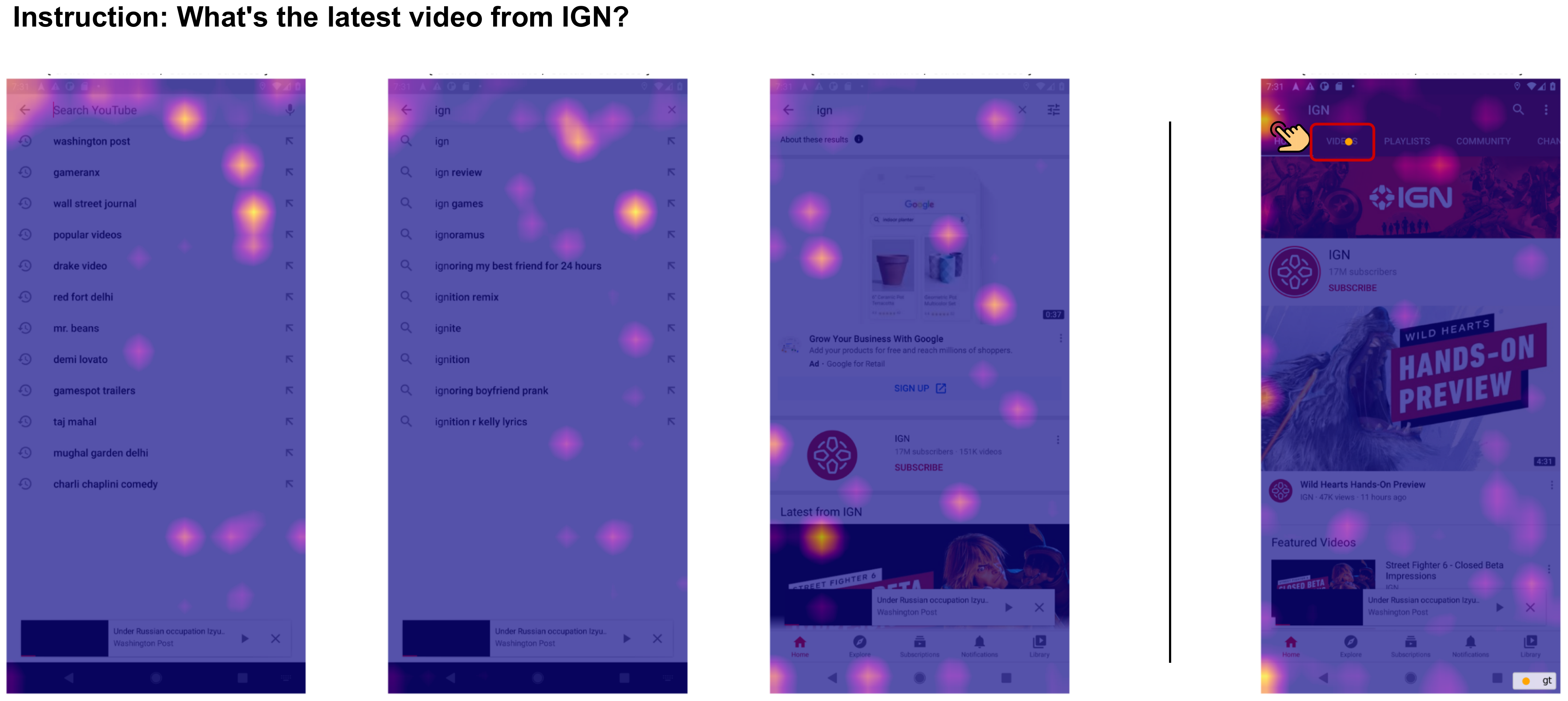} 
        \caption{Qwen2.5-VL 7B SFT 3AO}
    \end{subfigure}
    \caption{Use case: Attention visualization over the current image and three historical images within a single input.}
    \label{fig:three-graphs}
\end{figure*}

\vspace{0.5cm}

\begin{figure*}[ht]
    \centering
    \setlength{\abovecaptionskip}{0.01cm}
    \setlength{\belowcaptionskip}{0.01cm}
    \begin{subfigure}[b]{\textwidth}
        \centering
        \setlength{\abovecaptionskip}{0.01cm}
        \setlength{\belowcaptionskip}{0.01cm}
        \includegraphics[width=0.70\textwidth]{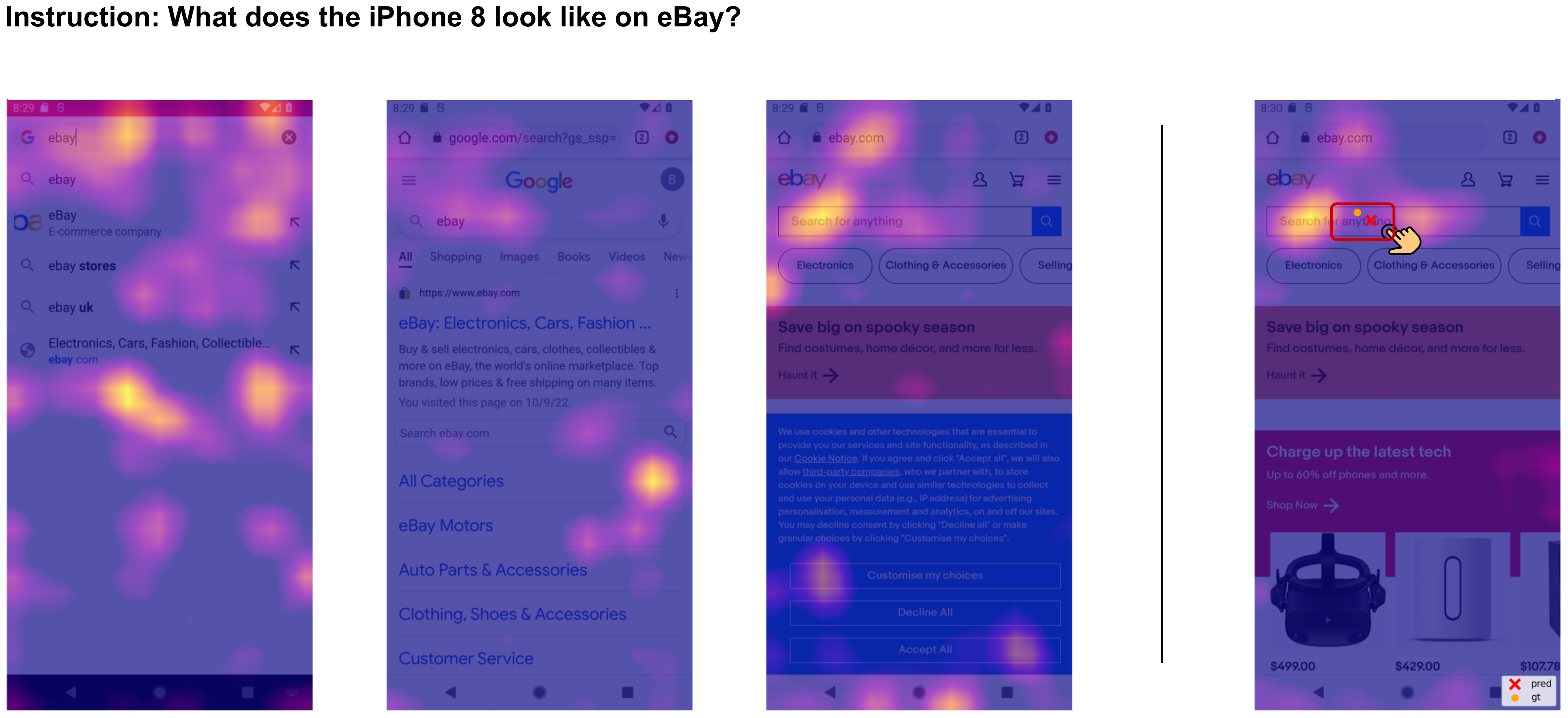} 
        \caption{CCPO 7B 3AO}
    \end{subfigure}
    \vspace{0.2cm}
    \begin{subfigure}[b]{\textwidth}
        \centering
        \setlength{\abovecaptionskip}{0.01cm}
        \setlength{\belowcaptionskip}{0.01cm}
        \includegraphics[width=0.70\textwidth]{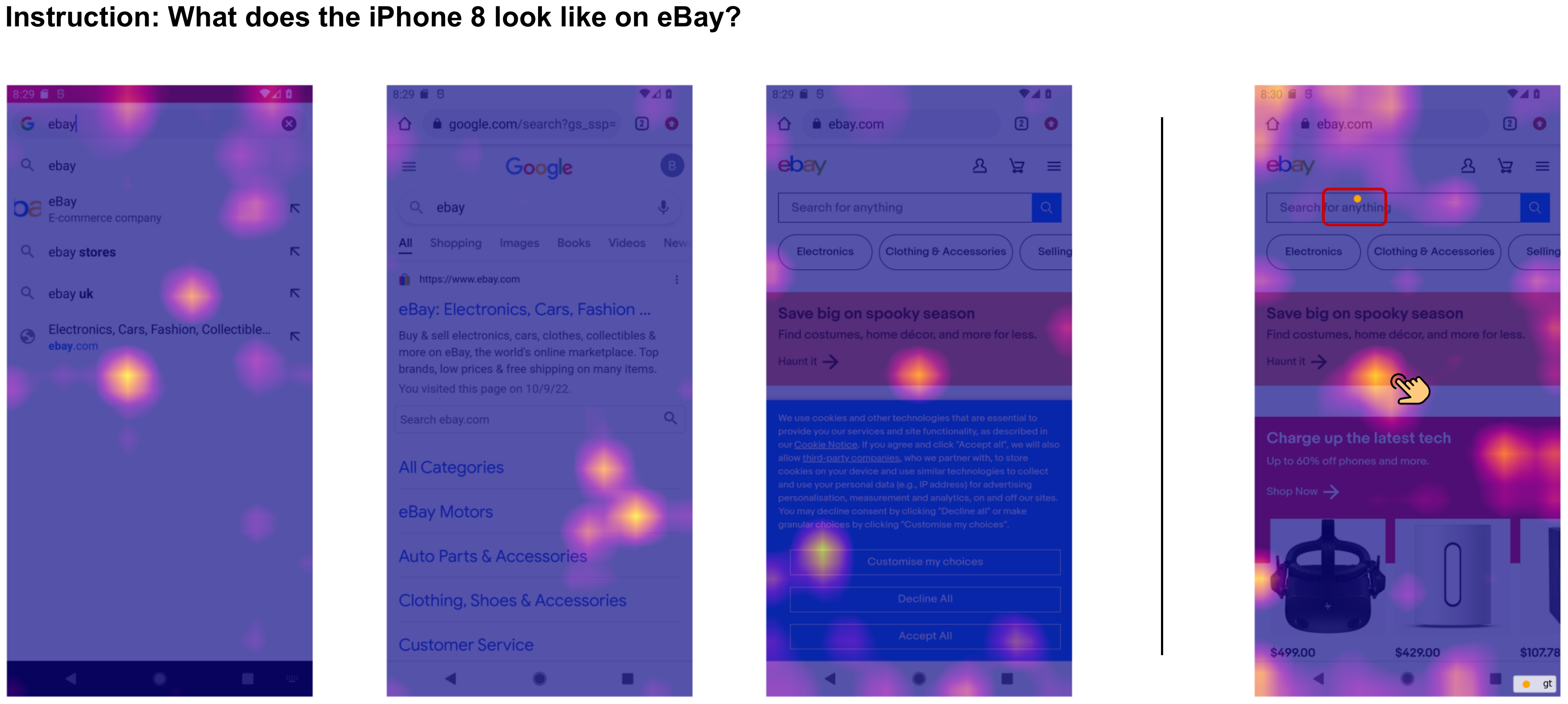} 
        \caption{Qwen2.5-VL 7B SFT 3AO}
    \end{subfigure}
    \caption{Use case: Attention visualization over the current image and three historical images within a single input.}
    \label{fig:three-graphs}
\end{figure*}

\begin{figure*}[ht]
    \centering
    \setlength{\abovecaptionskip}{0.01cm}
    \setlength{\belowcaptionskip}{0.01cm}
    \begin{subfigure}[b]{\textwidth}
        \centering
        \setlength{\abovecaptionskip}{0.01cm}
        \setlength{\belowcaptionskip}{0.01cm}
        \includegraphics[width=0.72\textwidth]{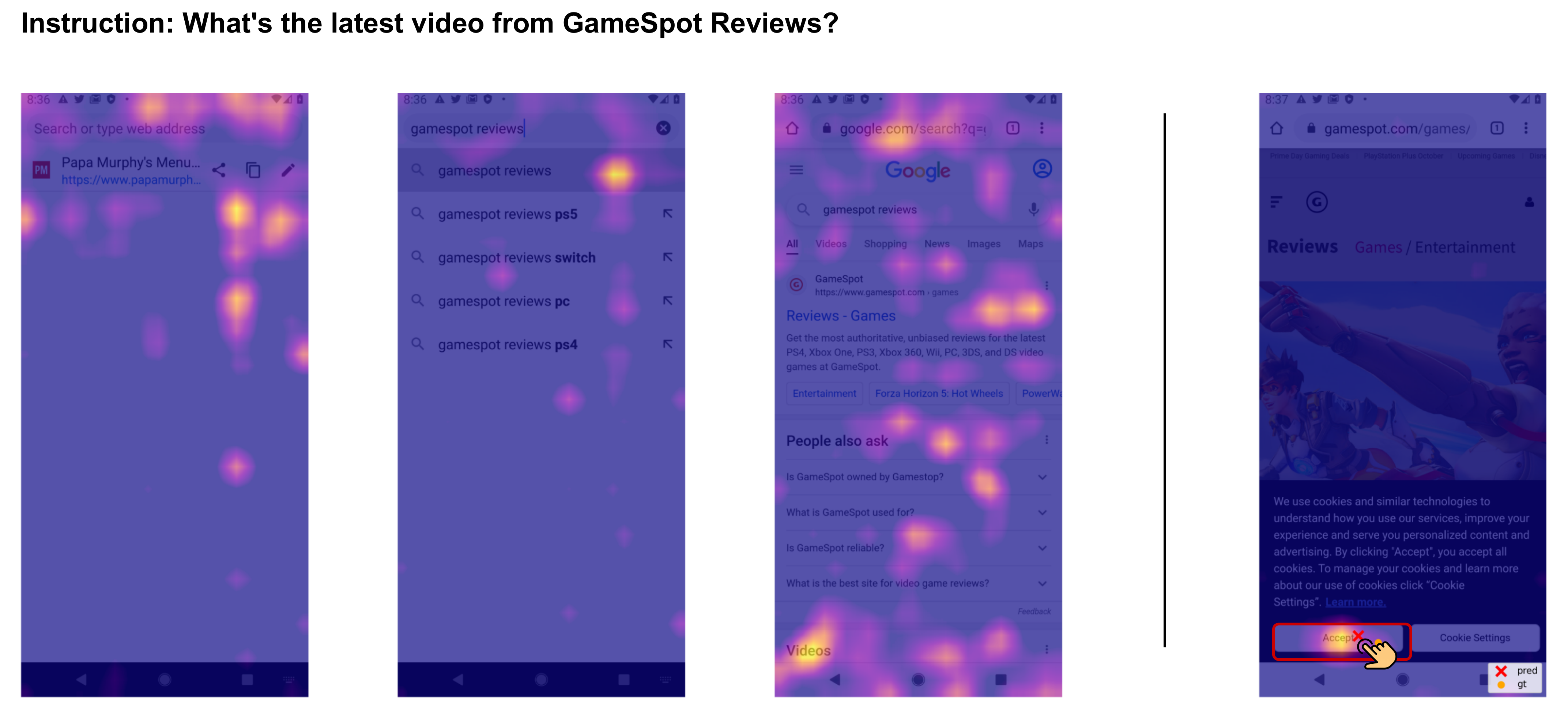} 
        \caption{CCPO 7B 3AO}
    \end{subfigure}
    \vspace{0.2cm}
    \begin{subfigure}[b]{\textwidth}
        \centering
        \setlength{\abovecaptionskip}{0.01cm}
        \setlength{\belowcaptionskip}{0.01cm}
        \includegraphics[width=0.72\textwidth]{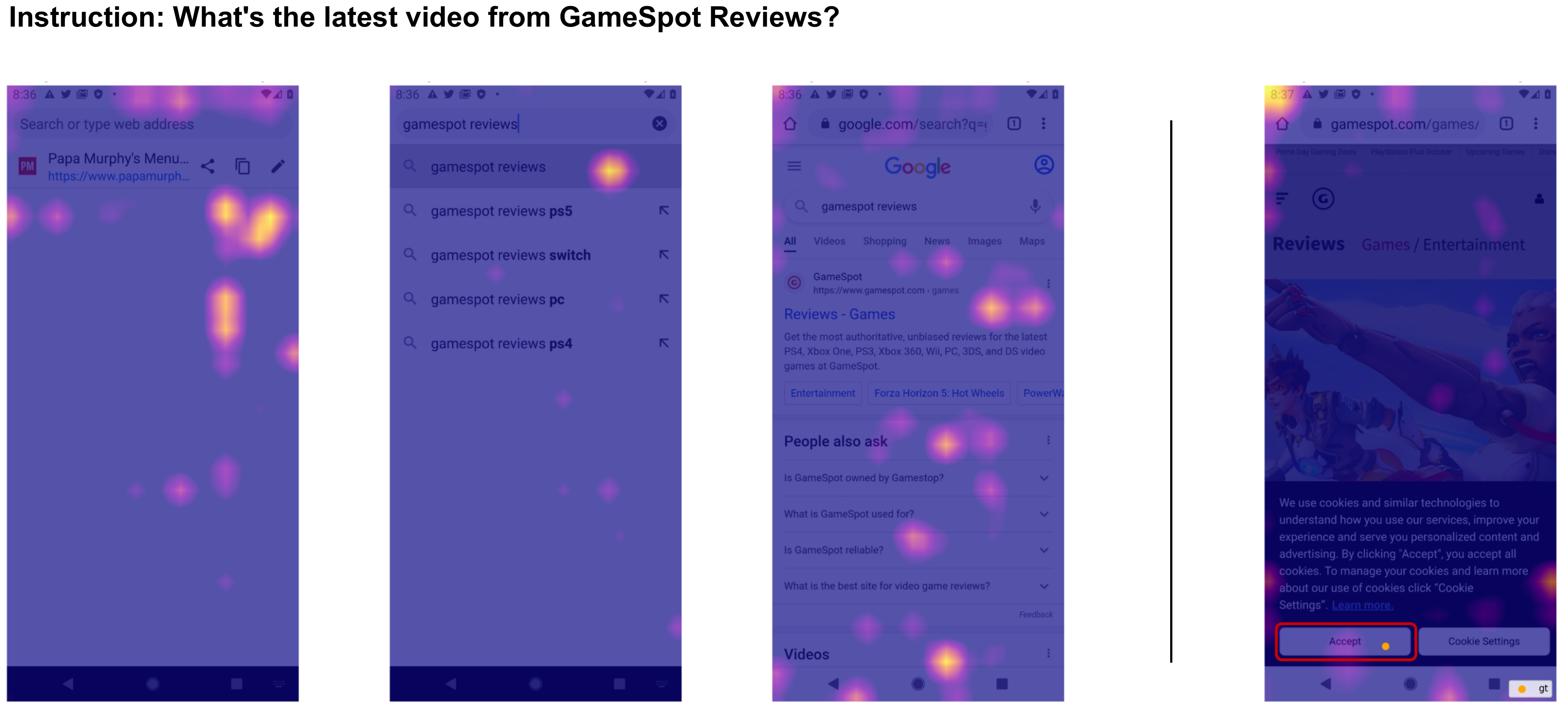} 
        \caption{Qwen2.5-VL 7B SFT 3AO}
    \end{subfigure}
    \caption{Use case: Attention visualization over the current image and three historical image within a single input.}
    \label{fig:three-graphs}
\end{figure*}

\begin{figure*}[ht]
    \centering
    \setlength{\abovecaptionskip}{0.01cm}
    \setlength{\belowcaptionskip}{0.01cm}
    \begin{subfigure}[b]{\textwidth}
        \centering
        \setlength{\abovecaptionskip}{0.01cm}
        \setlength{\belowcaptionskip}{0.01cm}
        \includegraphics[width=0.72\textwidth]{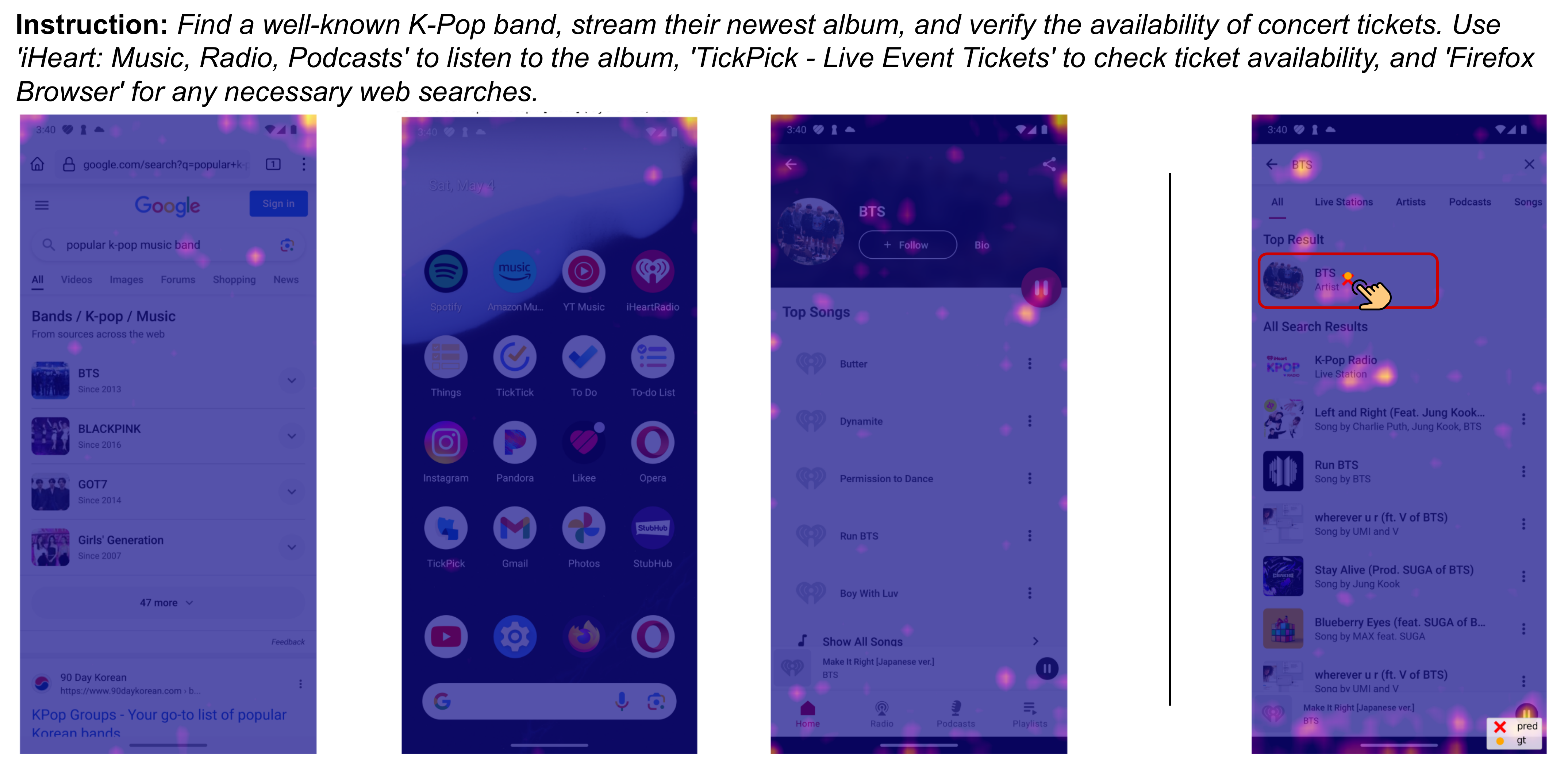} 
        \caption{CCPO 7B 3AO}
    \end{subfigure}
    \vspace{0.2cm}
    \begin{subfigure}[b]{\textwidth}
        \centering
        \setlength{\abovecaptionskip}{0.01cm}
        \setlength{\belowcaptionskip}{0.01cm}
        \includegraphics[width=0.72\textwidth]{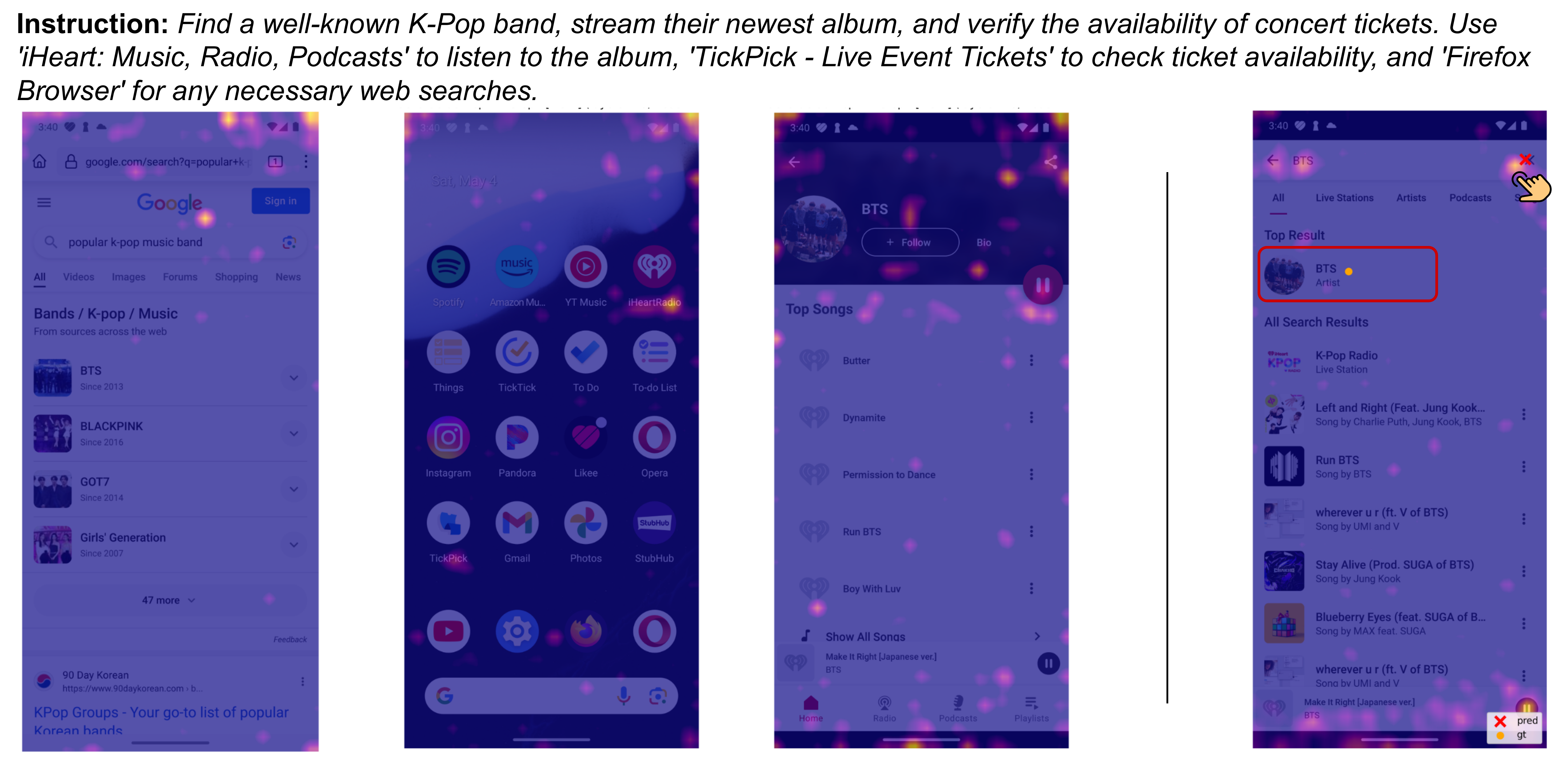} 
        \caption{Qwen2.5-VL 7B SFT 3AO}
    \end{subfigure}
    \caption{Use case: Attention visualization over the current image and three historical image within a single input.}
    \label{fig:three-graphs}
\end{figure*}

\begin{figure*}[ht]
  \centering
  \setlength{\abovecaptionskip}{0.01cm}
  \setlength{\belowcaptionskip}{0.01cm}
  \includegraphics[width=0.85\linewidth]{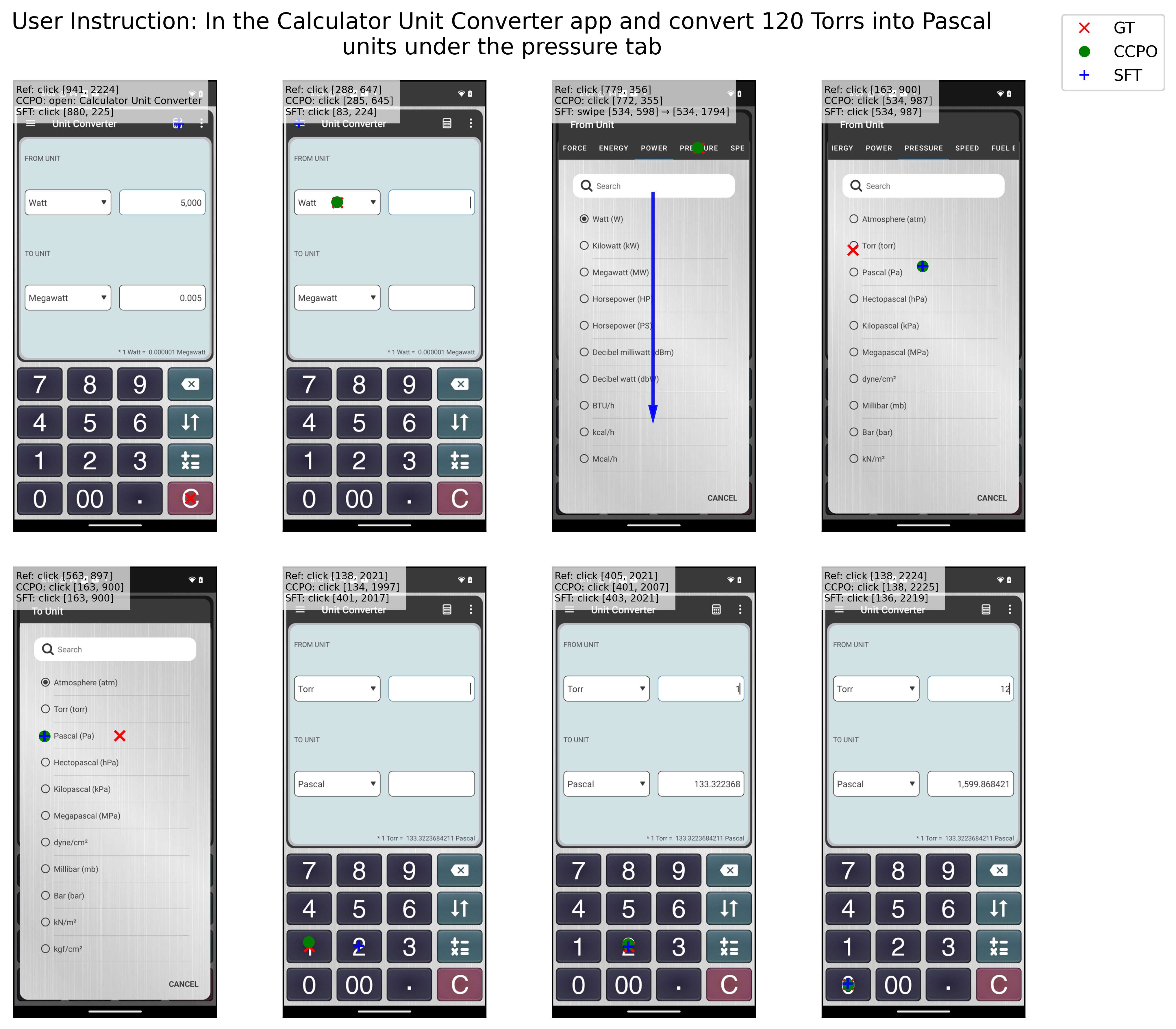} 
  \caption {Use case: Full-trajectory examples for CCPO and SFT.}
\end{figure*}


\begin{figure*}[ht]
  \centering
  \setlength{\abovecaptionskip}{0.01cm}
  \setlength{\belowcaptionskip}{0.01cm}
  \includegraphics[width=0.78\linewidth]{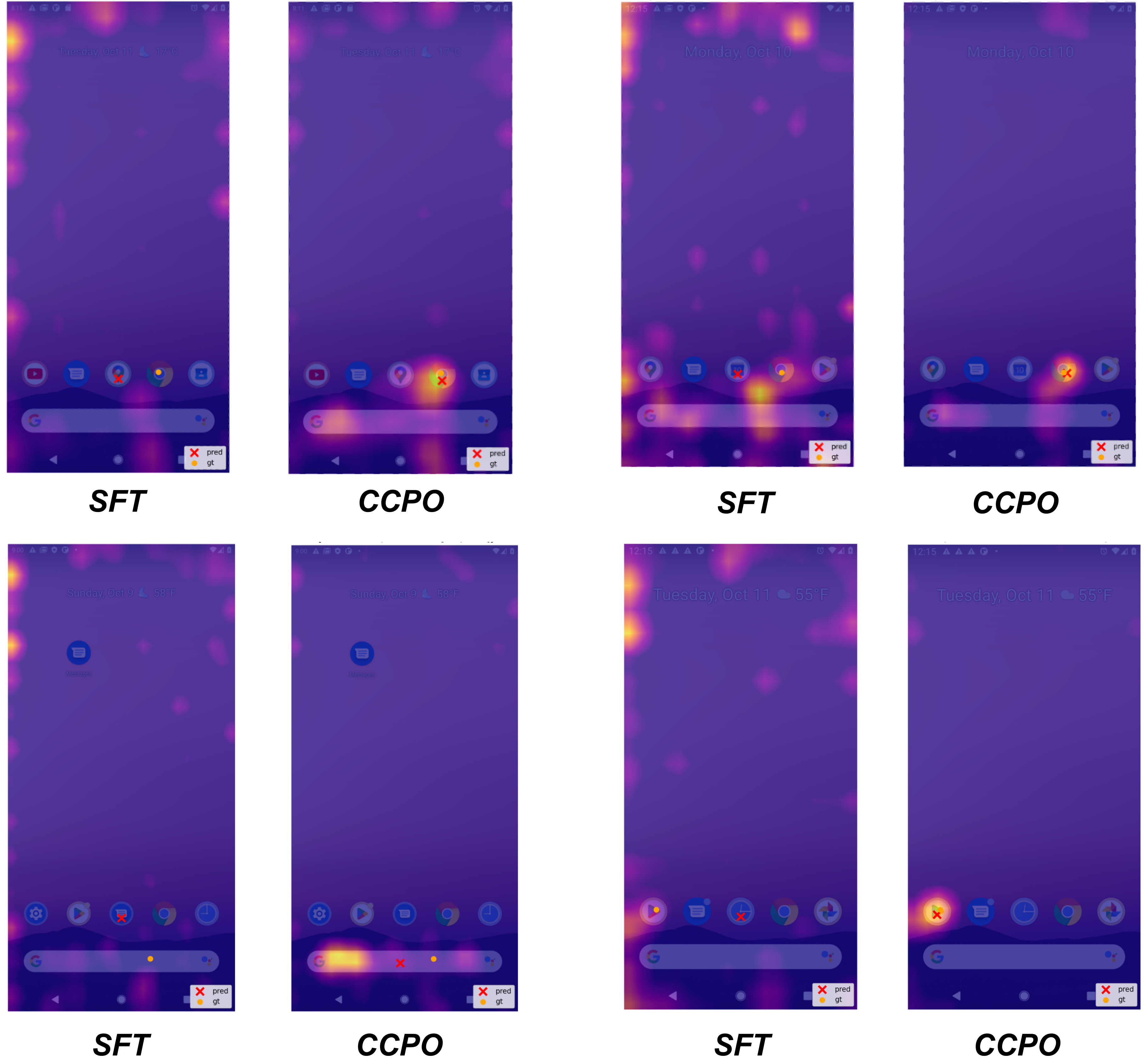} 
  \caption {Use case: CCPO demonstrates improved target-centric attention compared to SFT.}
\end{figure*}


\begin{figure*}[ht]
    \setlength{\abovecaptionskip}{0.01cm}
    \setlength{\belowcaptionskip}{0.01cm}
    \subsection{Failure Case}\addcontentsline{toc}{subsection}{Failure Case}
    \vspace{0.5cm}
    \centering
    \begin{subfigure}[b]{\textwidth}
        \centering
        \setlength{\abovecaptionskip}{0.01cm}
        \setlength{\belowcaptionskip}{0.01cm}
        \includegraphics[width=0.7\textwidth]{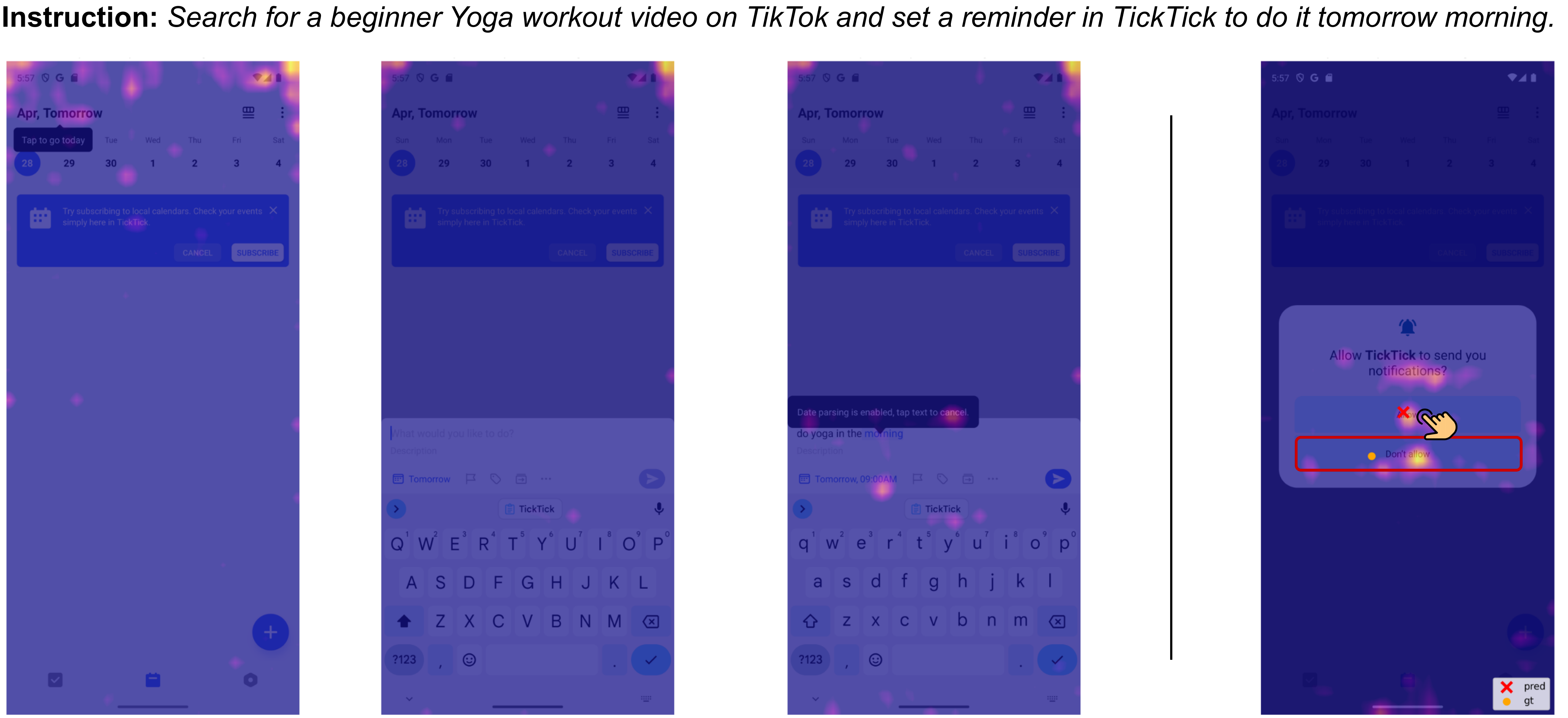} 
        \caption{CCPO 7B 3AO}
    \end{subfigure}
    \vspace{0.2cm}
    \begin{subfigure}[b]{\textwidth}
        \centering
        \setlength{\abovecaptionskip}{0.01cm}
        \setlength{\belowcaptionskip}{0.01cm}
        \includegraphics[width=0.7\textwidth]{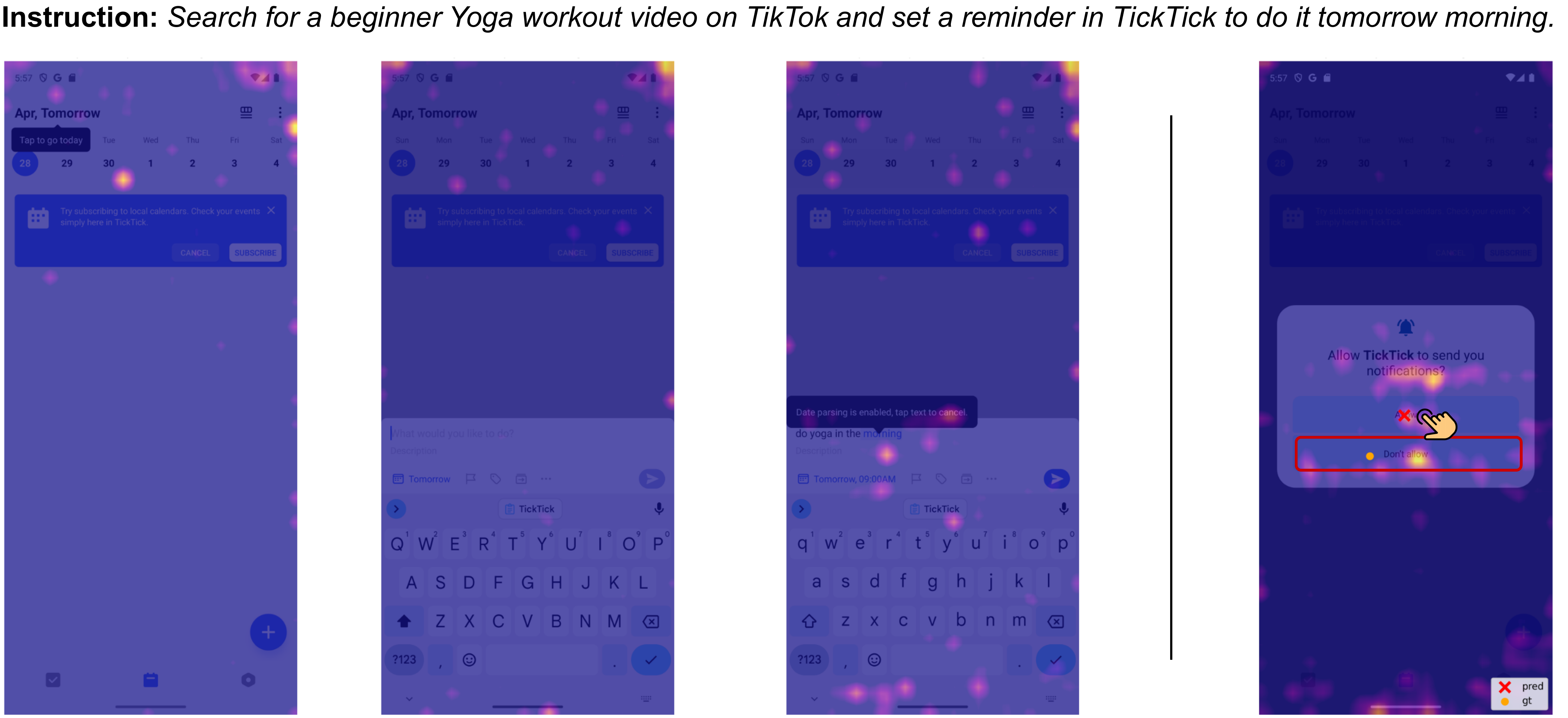} 
        \caption{Qwen2.5-VL 7B SFT 3AO}
    \end{subfigure}
    \caption{Failure case: Attention visualization over the current image and three historical images within a single input.}
    \label{fig:three-graphs}
\end{figure*}

\begin{figure*}[ht]
    \centering
    \setlength{\abovecaptionskip}{0.01cm}
    \setlength{\belowcaptionskip}{0.01cm}
    \begin{subfigure}[b]{\textwidth}
        \centering
        \setlength{\abovecaptionskip}{0.01cm}
        \setlength{\belowcaptionskip}{0.01cm}
        \includegraphics[width=0.7\textwidth]{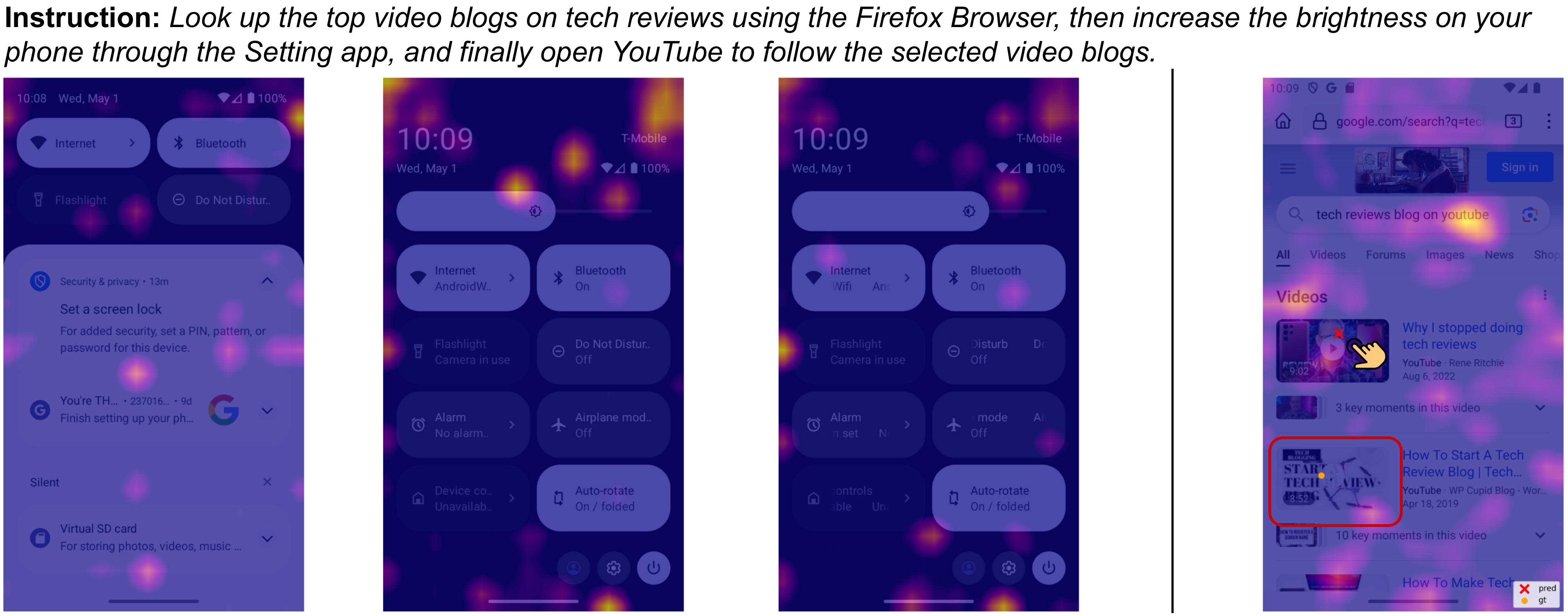} 
        \caption{CCPO 7B 3AO}
    \end{subfigure}
    \vspace{0.2cm}
    \begin{subfigure}[b]{\textwidth}
        \centering
        \setlength{\abovecaptionskip}{0.01cm}
        \setlength{\belowcaptionskip}{0.01cm}
        \includegraphics[width=0.7\textwidth]{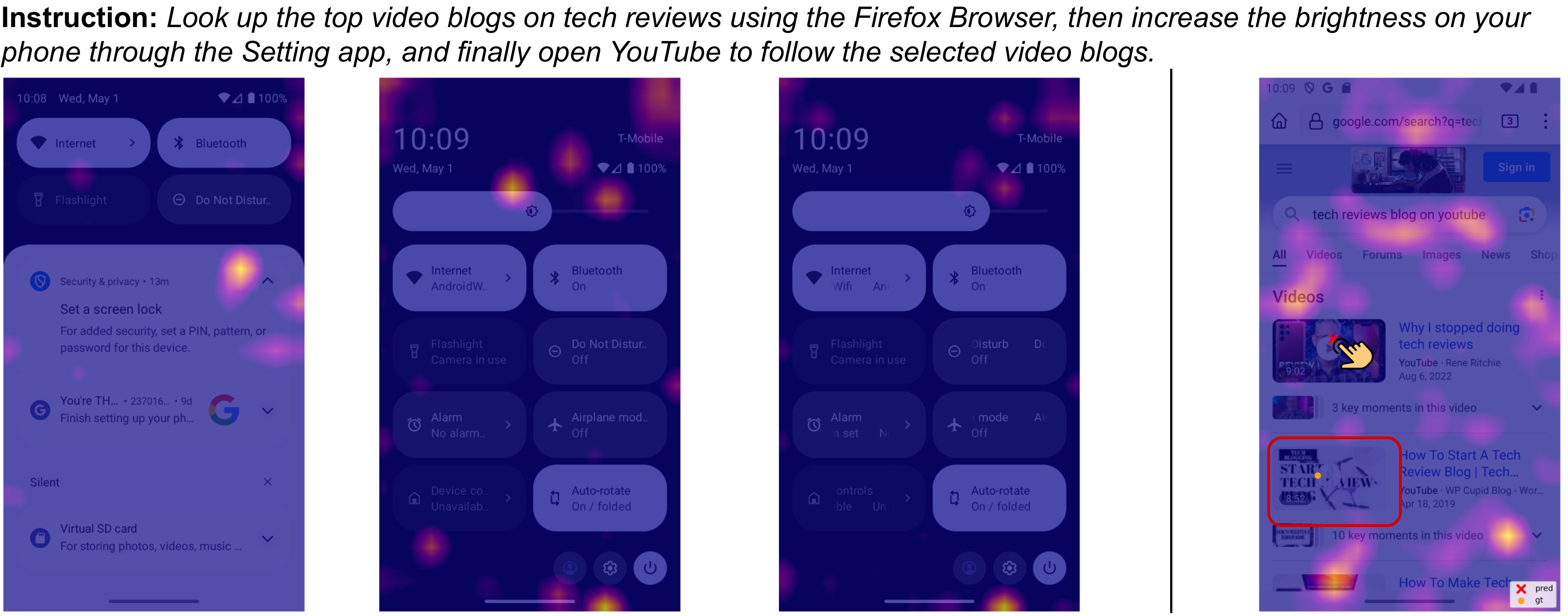} 
        \caption{Qwen2.5-VL 7B SFT 3AO}
    \end{subfigure}
    \caption{Failure case: Attention visualization over the current image and three historical image within a single input.}
    \label{fig:three-graphs}
\end{figure*}


\lstdefinestyle{promptstyle}{
  basicstyle=\ttfamily\footnotesize,
  columns=fullflexible,
  breaklines=true,
  breakatwhitespace=true,
  keepspaces=true,
  showstringspaces=false,
  frame=none
}

\begin{figure*}[t]
\subsection{Prompt}\addcontentsline{toc}{subsection}{Prompt}
\vspace{0.2cm}
\centering
\begin{tcolorbox}[enhanced,
  colback=gray!10, colframe=black,
  boxrule=1.2pt, arc=0pt,
  left=8pt,right=8pt,top=8pt,bottom=8pt,
  width=\textwidth
]
\begin{lstlisting}[style=promptstyle]
You are a GUI agent. You are given a task and your action history, with screenshots. You need to perform the next action to complete the task. 
## Output Format

```
<action> ... </action>
```

## Action Space

You can perform the following actions:
- key: Perform a key event on the mobile device using adb's `keyevent` syntax.
- click: Click the point on the screen with specified (x, y) coordinates.
- long_press: Press the point on the screen with specified (x, y) coordinates for a specified number of seconds.
- swipe: Swipe from starting point with specified (x, y) coordinates to endpoint with specified (x2, y2) coordinates.
- type: Input the specified text into the activated input box.
- answer: Output the specified answer.
- system_button: Press the specified system button: Back, Home, Menu, or Enter.
- open: Open an application on the device specified by text.
- wait: Wait for a specified number of seconds for changes to occur.
- terminate: Terminate the current task and report its completion status: success or failure.

The arguments you can use are:
- coordinate: (x, y): The x and y pixels coordinates from the left and top edges.
- coordinate2: (x, y): The x and y pixels coordinates from the left and top edges for the endpoint of a swipe.
- text: Text input required by actions like `key`, `type`, `answer`, and `open`.
- time: The time in seconds required by actions like `long_press` and `wait`.
- button: System buttons available for pressing: Back, Home, Menu, or Enter. Possible values: Back, Home, Menu, Enter.
- status: The completion status of a terminated task. Possible values: success, failure.

Format your output as a JSON object with the selected action and its arguments at the same level.

Example outputs:
<action>
{"action": "key", "text": "<value>"}
</action>
<action>
{"action": "click", "coordinate": "<value>"}
</action>

## Note

- Planing the task and explain your reasoning step-by-step in `think` part.
- Write your action in the `action` part according to the action space.
\end{lstlisting}
\end{tcolorbox}
\end{figure*}